%% file: main.tex
\documentclass[onecolumn]{cohere}

\usepackage{lipsum}

\input{math_commands.tex}

\usepackage{stfloats} 
\usepackage{hyperref}
\usepackage{url}
\usepackage{booktabs}
\usepackage{graphicx}
\usepackage{multirow}
\usepackage[export]{adjustbox}
\usepackage{float}
\usepackage{caption}
\usepackage{subcaption}
\usepackage[colorinlistoftodos]{todonotes}
\usepackage{lscape}
\usepackage{rotating}
\usepackage{enumitem}
\usepackage{wrapfig}
\usepackage{tabularx}
\usepackage{xcolor}
\usepackage{tcolorbox}
\usepackage{mdframed}
\usepackage{booktabs} 
\usepackage{siunitx}  
\tcbuselibrary{most}
\usepackage{colortbl}

\definecolor{cblue}{rgb}{0.39, 0.58, 0.92}
\definecolor{rbrown}{rgb}{0.73, 0.56, 0.56}
\definecolor{salmon}{rgb}{0.98, 0.50, 0.45}
\definecolor{darkgray1}{rgb}{0.6627, 0.6627, 0.6627}
\definecolor{greenish}{rgb}{0.9, 0.98, 0.9}
\definecolor{reddish}{rgb}{0.98, 0.9, 0.9} 
\setcitestyle{numbers,square}

\newtcolorbox{promptbox}{
  colback=cblue!5!white,
  colframe=cblue!75!black,
  title=Prompt,
  fonttitle=\bfseries,
  coltitle=white,
  colbacktitle=cblue!75!black,
  enhanced,
  arc=2mm,
  boxrule=1.5pt,
  drop shadow
}

\newtcolorbox{completionbox}[1]{
  colback=rbrown!5!white,
  colframe=rbrown!75!black,
  title=Completion #1,
  fonttitle=\bfseries,
  coltitle=white,
  colbacktitle=rbrown!75!black,
  enhanced,
  arc=2mm,
  boxrule=1.5pt,
  drop shadow
}

\newtcolorbox{greenbox}{
  colback=greenish,
  colframe=greenish,
  sharp corners,
  boxrule=1pt,
  enhanced,
  width=0.95\linewidth,
  boxsep=0pt,
  left=1pt,
  right=1pt,
}

\newtcolorbox{redbox}{
  colback=reddish,
  colframe=reddish,
  sharp corners,
  boxrule=1pt,
  enhanced,
  width=0.95\linewidth,
  boxsep=0pt,
  left=1pt,
  right=1pt,
}

\newtcolorbox{boxA}{
  colback=cblue!5,
  colframe=cblue,
  left=3pt,
  right=3pt,
  boxsep=2pt
}

\newtcolorbox{boxB}{
  colback=rbrown!5,
  colframe=rbrown,
  left=3pt,
  right=3pt,
  boxsep=2pt
}

\newtcolorbox{boxTie}{
  colback=salmon!5,
  colframe=salmon,
  left=3pt,
  right=3pt,
  boxsep=2pt
}

\title{Which Prompts Make The Difference? \\Data Prioritization For Efficient Human \\LLM Evaluation}

\author{
    name={Meriem Boubdir},
    affiliation={Cohere for AI},
    email={meri.boubdir@gmail.com}
}
\author{
    name={Edward Kim},
    affiliation={Cohere},
    email={edward@cohere.com}
}
\author{
    name={Beyza Ermis},
    affiliation={Cohere for AI},
    email={beyza@cohere.com}
}
\author{
    name={Marzieh Fadaee},
    affiliation={Cohere for AI},
    email={marzieh@cohere.com}
}
\author{
    name={Sara Hooker},
    affiliation={Cohere for AI},
    email={sarahooker@cohere.com}
}

\date{\today}
\abstract{
Human evaluation is increasingly critical for assessing large language models, capturing linguistic nuances, and reflecting user preferences more accurately than traditional automated metrics. However, the resource-intensive nature of this type of annotation process poses significant challenges. The key question driving our work: \textit{is it feasible to minimize human-in-the-loop feedback by prioritizing data instances which most effectively distinguish between models?} We evaluate several metric-based methods and find that these metrics enhance the efficiency of human evaluations by minimizing the number of required annotations, thus saving time and cost, while ensuring a robust performance evaluation. We show that our method is effective across widely used model families, reducing instances of indecisive (or ``tie'') outcomes by up to $\text{54\%}$ compared to a random sample when focusing on the top-20 percentile of prioritized instances. This potential reduction in required human effort positions our approach as a valuable strategy in future large language model evaluations.
}

\begin{document}
\sloppy

\section{Introduction}
\label{sec:intro}

Large language models (LLMs) have produced notable breakthroughs in downstream performance~\citep{raffel2020exploring,brown2020language,chowdhery2022palm,roy2022benchclamp,zhu2023multilingual,maynez-etal-2023-benchmarking,biderman2023pythia,touvron2023llama}, but have also introduced new challenges in model evaluation. The success of LLMs has initiated a fundamental paradigm shift away from small specialized models designed for single tasks to \textit{universal} models expected to perform well across a wide range of tasks.
This shift has also posed an existential challenge for evaluation, with a need to move away from solely task-specific automatic metrics of evaluation and increasing reliance on human evaluation. 

While automatic metrics offer a degree of objectivity and reproducibility, alongside the benefits of speed and cost-effectiveness, they often fall short in fully capturing the complexities and nuances of natural language~\citep{mathur-etal-2020-tangled,shen-etal-2022-evaluation}.
Moreover, automatic metrics often rely on auxiliary models which introduce potential points of failure and unexpected challenges over time~\citep{pozzobon2023challenges}. For example, reference-based metrics such as BLEU~\citep{papineni-etal-2002-bleu} and ROUGE~\citep{lin-2004-rouge} are usually poor indicators of human judgment, as they emphasize lexical overlap and struggle to account for the diverse expressions inherent in semantic representation ~\citep{Graham2015ReevaluatingAS,Yang2018AdaptationsOR,Blagec2022AGA}.
This limitation is particularly prominent in open-ended conversations \citep{liu-etal-2016-evaluate}, open-response question answering \citep{kocisky-etal-2018-narrativeqa}, summarization \citep{schluter-2017-limits}, and code generation \citep{Evtikhiev2022OutOT}, where there isn't a single reference ground truth for comparison, but rather diverse yet equivalent responses.

Human evaluation has emerged as an indispensable method in assessing LLMs' performance. It provides direct feedback on understandability, usability, correctness, and safety \citep{ung-etal-2022-saferdialogues} from the end-user's perspective.
Furthermore, human feedback plays an essential role in aligning model behavior through either supervised finetuning on human annotated completions \citep{bakker2022finetuning,rafailov2023direct,zhao2023slichf} or reinforcement learning with reward models learned from human preferences ~\citep{ziegler2020finetuning,stiennon2020learning,bai2022training}.

However, this growing dependence on human evaluation has introduced certain trade-offs. The evaluation process itself is resource-intensive requiring substantial investments in human resources and domain-specific expertise~\citep{cohen-etal-2018-three,beck-etal-2020-representation}. Notably, in the context of safety, human annotation can expose evaluators to highly toxic content, which carries a significant risk to their mental health and leaves them vulnerable to lasting psychological harm \citep{steiger2021psychological}.

\begin{figure}[t]
    \centering
    \includegraphics[width=0.9\linewidth]{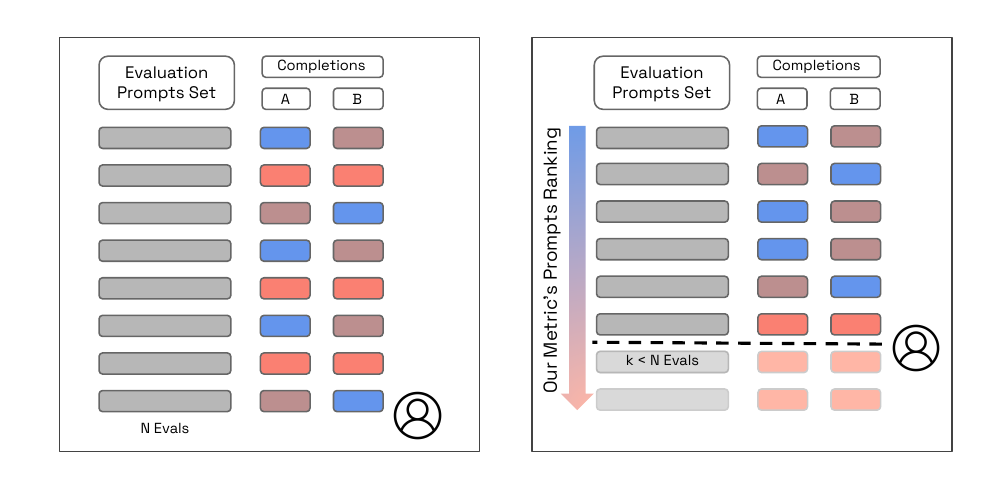}
    \caption{\textbf{Evaluation Methodology}: Human evaluators assess N prompts, each with completions from two different models, A and B (left). Completions are ranked by dissimilarity, from the most distinct to potential ties (right). The dashed line marks a threshold at `k' evaluations beyond which the likelihood of a tied outcome increases, making further evaluations less informative at assessing models performance discrepancies.}
    \label{fig:methodology}
\end{figure}

In this work, we aim to reduce the number of required human annotations while mitigating tie outcomes in a comparison setting, thereby diminishing both time and resource expenditures. We ask \textit{Which prompts should be prioritized to most efficiently differentiate and rank relative model quality amongst a pool of models?} To achieve this, we introduce prompt prioritization as a strategy to reduce the need for extensive human feedback and effectively decrease the time and cost complexities associated with the annotation collection process. By informatively mitigating tie outcomes in a comparison setting, our aim is to maintain robust performance results. 

We benchmark several metrics, such as Cross-Entropy and KL divergence, to assess the dissimilarity between model completions.
These metrics, efficient to compute as derivatives of a model's generation probabilities, guide our prompt ranking strategy. Across several popular and widely used model families such as Flan-T5 \citep{https://doi.org/10.48550/arxiv.2210.11416}, Dolly-V2 \citep{DatabricksBlog2023DollyV2} families, as well as MPT-7B-instruct \citep{MosaicML2023Introducing} and Falcon-7B-instruct \citep{refinedweb}, we arrive at consistent results. Our contributions are as follows:
\begin{enumerate}
\itemsep0em 
    \item \textbf{Systematic Offline Ranking}: We introduce a systematic, offline ranking method to expedite the evaluation process. 
    This approach leverages KL divergence and Cross-Entropy to prioritize prompts and completion pairs based on their predicted potential to yield decisive preference outcomes.
    \item \textbf{Reduction in Ties}: We show that our use of KL divergence for ranking completion pairs results in a remarkable $\text{54\%}$ reduction in ties within the top 20\% of prioritized prompts when compared to random selection. This finding highlights the potential of our approach in improving the efficiency of human-in-the-loop evaluations.  
    \item \textbf{Robustness of Elo Scores}: We highlight that our ranking strategy enhances the stability of Elo scores, a widely used evaluation metric for players in zero-sum games, across various model comparisons. This improvement allows us to reduce our dependence on extensive human annotations while maintaining reliable and conclusive models rankings.
\end{enumerate}

\input{methods}

\input{experiment}

\input{results}

\input{related}

\section{Conclusion}
\label{sec:conc}
This work was motivated by the problem of language model preference evaluation, with the primary objective of improving the efficiency of the manual annotation process. We proposed a data prioritization approach that identifies prompts in the evaluation set with minimal information gain when annotated. Through the early elimination of prompts resulting in tied outcomes during the annotation process, we acquire robust signals regarding model preference, enabling us to determine such preferences with fewer annotations. Our ranking method achieved an impressive reduction of up to $54\%$ in tied outcomes among the top-20 percentile of the evaluation set compared to random ranking. Moreover, when evaluating subsets of our prioritized data, we found that Elo scores, a measure of model performance, showed clearer distinctions between models in the early thresholds like the top 20\% to 30\%. These results are consistent with our `Gold Standard' models ranking, underscoring the efficacy of our method over random sequencing, especially in resource-limited scenarios. Our findings have a crucial impact on the manual preference evaluation of language models, and carries broader implications for increasing the overall efficiency of the human annotation process.

\bibliography{main}

\appendix
\input{appendix}

\end{document}

%% file: math_commands.tex
\usepackage{amsmath,amsfonts,bm}

\def\eqref#1{equation~\ref{#1}}

\def\1{\bm{1}}

\DeclareMathAlphabet{\mathsfit}{\encodingdefault}{\sfdefault}{m}{sl}
\SetMathAlphabet{\mathsfit}{bold}{\encodingdefault}{\sfdefault}{bx}{n}



%% file: methods.tex
\section{Methodology}
\label{sec:methodology}

Many human evaluation pipelines use a pairwise comparison format where model A is compared to model B.
Pairwise evaluation is often considered to be less subjective and more consistent compared to other evaluation methods because it focuses on relative judgments rather than absolute judgments.
This involves comparing the responses of two models to a single prompt, a strategy referred to as a head-to-head pairwise block comparison, dating back to the work of ~\citet{Thurstone1927-THUALO-2}.  
When a comparison for a given prompt favors one completion over the other, we gain insight into the relative performance of the two models (e.g., Sample 1 in Figure~\ref{fig:prompt-samples}).
If both models yield high-quality and equally valid responses for a prompt, discerning the superiority of one over the other becomes inherently challenging, as demonstrated in Figure~\ref{fig:prompt-samples} (Sample 2). 
In this work, our objective involves strategically selecting prompts that amplify the informativeness of each comparison, thereby streamlining and optimizing the evaluation process.

\begin{figure}[t]
    \centering
    \begin{minipage}[t]{0.48\linewidth}
        \begin{greenbox}
          \textbf{Sample 1: Comparison breaks tie -- effective for ranking models} \\
          \textit{Clear preference for completion A. This comparison has low likelihood of ties.}\\[.5em]
          \begin{promptbox}
            Put the concepts together to form a sentence: kitchen, sit, table.
          \end{promptbox}
        
          \begin{completionbox}{A}
            A man sits at a table in a kitchen.
          \end{completionbox}
        
          \begin{completionbox}{B}
            The kitchen sits at the table.
          \end{completionbox}
          \hfill
        \end{greenbox}
    \end{minipage}%
    \begin{minipage}[t]{0.48\linewidth}
        \begin{redbox}
          \textbf{Sample 2: Comparison results in a tie -- not effective for ranking models} \\
          \textit{Both completions present equivalently suitable coherent and rational responses.}\\[.5em]
          \begin{promptbox}
            Put the concepts together to form a sentence: jump, pool, swimsuit.
          \end{promptbox}
        
          \begin{completionbox}{A}
            A girl in a swimsuit jumps into a pool.
          \end{completionbox}
        
          \begin{completionbox}{B}
            Jumping into a swimming pool without a swimsuit is a bad idea.
          \end{completionbox}
        \end{redbox}
    \end{minipage}
    \caption{Comparison of a prompt, used to rank completions from models A and B, \textbf{Left:} which is strong at distinguishing which model is better (low likelihood of ties in human eval), \textbf{Right:} which is weak at distinguishing which model is better (high likelihood of ties in human eval). Our goal is to automatically identify and rank prompts which are similar to sample 1 higher than the ones similar to sample 2.}
    \label{fig:prompt-samples}
\end{figure}

Formally, let us consider an evaluation set $P = \{p_i\}_{i=1}^N$ where $p_i$ represents a prompt.
For every prompt $p_i$ within $P$, and for every model $M$ within our pool of models $\mathbb{M}$, a completion $c_{i}^{M}$ is generated.
Paired model completions are formed for evaluation, denoted as $C = \{(c_i^{A}, c_i^{B})\}_{i=1}^N$, where $A$ and $B$ are two distinct models from our model pool $\mathbb{M}$.
Subsequently, an annotator reviews each pair to assess their relative quality.
The outcomes of these human evaluations are captured as evaluation scores $(Score_i^A, Score_i^B)$, defined as:
\begin{equation}
    \begin{cases}
    Score_i^A = 1 \text{ and } Score_i^B = 0 & \text{if $c_i^A$ is preferred}\\
    Score_i^A = 0 \text{ and } Score_i^B = 1  & \text{if $c_i^B$ is preferred}\\
    Score_i^A = Score_i^B = 1  & \text{if both are of similar quality} \\
    \end{cases}
\label{eq:eval_score}
\end{equation}

We propose an offline approach that automatically ranks prompts in $P$ using model completions.
While we assess responses from two models, our primary focus is on highlighting their dissimilarity.
Although widely used, conventional string matching techniques such as BLEU and ROUGE are not well-suited for this problem. These metrics may indicate a high lexical overlap between two completions, but they might differ significantly in meaning or quality.
Coversely, completions may exhibit minimal lexical overlap while delivering equally valid answers to the same prompt (see Figure~\ref{fig:prompt-samples}).
Such a focus is instrumental in minimizing tie outcomes in human evaluations, which can arise when both completions are viewed by annotators as similarly good or bad.

Our goal is to rearrange the prompts and present them to the annotators in an order that starts with a low likelihood of a tie outcome and progresses to a higher likelihood.
To achieve this, we reorder the prompt and completion pairs within an evaluation set $P$ by finding an optimal permutation $\pi$ based on dissimilarity scores to create the ordered set $\hat{P_\pi} = \{p_\pi(i)\}_{i=1}^N$.
This reordering aims to prioritize evaluation instances that yield a strong preference signal from the annotators.
Unlike the original set $P$, where an annotator would need to review all $N$ samples to conclude model preference, using the ordered set $\hat{P}$ will only require up to $k$ annotations, where $k < N$ (see Figure~\ref{fig:methodology}).

\subsection{Quantifying the "A vs. B" dissimilarity}

To enhance our evaluation process effectively, it is essential to understand how two model outputs diverge when given the same prompt.
This section outlines our strategy to systematically prioritize prompts that tend to diminish tie outcomes in the top percentile, while simultaneously maximizing them in the bottom percentile.

In this context, we quantify the variation between the responses of different models $\{c_i^{M_1}, ..., c_i^{M_n}\}$ for a given input prompt $p_i$, where $n$ is the number of models within our evaluation pool. 

To achieve this, we investigate scoring metrics that utilize the log probabilities or probabilities of the completions, as computed by their respective generating models. This serves as an efficient proxy metric for quality and has been demonstrated to be more calibrated as an indicator of model certainty at larger model sizes \citep{chen2023close}.

Let $S^A = (s_1^A, s_2^A, \ldots, s_t^A)$ and $S^B = (s_1^B, s_2^B, \ldots, s_t^B)$ be the log probabilities of tokens in the completions $c^A$ and $c^B$ respectively, where $t$ is the maximum length of the completions, accounting for padding. We compute the sequence probabilities $\mathbf{p}^A = (\exp(s_j^A))_{j=1}^{t}$ and $\mathbf{p}^B = (\exp(s_j^B))_{j=1}^{t}$.
To accomplish our objective, we employ both \textit{KL Divergence} (Equation~\ref{eq:kl}) and \textit{Cross-Entropy} (Equation~\ref{eq:ce}) as thesemetrics quantify the difference between two probability distributions.

\begin{equation}
    KL=\sum_i \mathbf{p}_i^A \log \frac{\mathbf{p}_i^A}{\mathbf{p}_i^B}
    \label{eq:kl}
\end{equation}
\begin{equation}
    CE = - \sum_i \mathbf{p}_i^A \log \mathbf{p}_i^B
    \label{eq:ce}
\end{equation}

\textit{KL Divergence} measures how one probability distribution diverges from a second, expected distribution, making it useful for capturing differences between model completions, $S^A$ and $S^B$. We normalize each sequence by the sum of its probabilities before calculating the divergence. \textit{Cross-Entropy} indicates the dissimilarity between the two distributions, serving as an indicator of the model's relative performance.

Once we have model completions for each model in $\mathbb{M}$ (representing our evaluation pool of models), we proceed to compute the values of our proposed metrics for each pair of completions.
We subsequently rank our prompt candidates according to each metric (KL Divergence and Cross-Entropy), resulting in ordered sets $\hat{P}_{CE}$ and $\hat{P}_{KL}$ respectively.
The prompts in each ordered set span from pairs with the most dissimilar completions to those with the least dissimilar completions according to the metric.
We anticipate that this prompt ranking will lead to a more efficient differentiation of model performance during human evaluation.

\subsection{Applying our ranking in Elo rating computation}
\label{sec:elo}

Originally introduced by ~\citet{elo1978rating} for ranking chess players, the Elo rating system has emerged as a pivotal tool in evaluating and comparing models performance, particularly in pairwise preference-based setups ~\citep{stiennon2020learning,askell2021general,menick2022teaching,bai2022training,nakano2022webgpt}.
By computing Elo ratings for each model within a pool, we can establish a global ranking tier based solely on pairwise comparisons.
In this section, we explore how our proposed evaluation methodology can benefit the convergence and robustness of Elo-based evaluations, especially when dealing with limited resources.

The appeal of the Elo rating system in evaluating LLMs lies in its ability to effectively capture the performance dynamics within a constantly changing pool of models.
Furthermore, it provides an intuitive interpretation of the ratings as a measure of expected performance in a direct head-to-head comparison of any two models.

By definition, the expected Elo scores for each model are calculated as:

\begin{subequations}
\begin{align}
    E_\mathsf{A} &= \frac{1}{1 + 10^{\frac{R_\mathsf{B} - R_\mathsf{A}}{400}}} \label{eq:elo-expa} \\
    E_\mathsf{B} &= \frac{1}{1 + 10^{\frac{R_\mathsf{A} - R_\mathsf{B}}{400}}} \label{eq:elo-expb}
\end{align}
\end{subequations}

Here $R_\mathsf{A}$ and $R_\mathsf{B}$ represent the ratings of models $\mathsf{A}$ and $\mathsf{B}$, respectively.
Following an evaluation instance, a model's rating is updated based on both its prior rating and its performance relative to expectations. 
The updated rating for model $\mathsf{A}$ is given by:
\begin{equation}
    R_{\mathsf{A}}'=R_{\mathsf{A}}+K \cdot (S_{\mathsf{A}}-E_{\mathsf{A}})
    \label{eq:elo-iter}
\end{equation}
In this equation, $R_{\mathsf{A}}$ is the current rating of model $\mathsf{A}$ and $E_\mathsf{A}$ its expected score. 
The term $K$ denotes the the impact magnitude of the evaluation outcome on the rating update.
The actual score of model $\mathsf{A}$, represented by $S_\mathsf{A}$, can be 1 for a win, 1/2 for a tie or 0 for a loss, as given by equation \ref{eq:eval_score}. Similarly, the rating update mechanism is applied to model $\mathsf{B}$.

After $N$ evaluations, the final Elo ratings, referred to as Elo scores are used to rank models.
Higher scores signify stronger models, and the scores difference between models serves as a valuable indicator of performance disparities.
As indicated in Equation~\ref{eq:elo-iter}, Elo ratings are updated through pairwise comparisons. 
The arrangement of evaluation prompts and, subsequently, the outcome of each comparison, can influence the behavior of the Elo ratings.
We use this evaluation framework to investigate how our prompt rankings impact the robustness of Elo scores and consequently model performance rankings.

%% file: experiment.tex
\section{Experimental Setup}
\label{sec:experiments}

\subsection{Pool of prompts} 
Our objective is to evaluate model quality across a range of tasks, spanning from unstructured to more structured formats. 
To achieve this, we sample prompts from several different sources. 
The social \emph{chit-chat} dataset, \textsc{Soda} \citep{kim2022soda}, is utilized to assess the models' performance in handling informal conversations.
To evaluate the models' proficiency in structured and goal-oriented settings, we incorporate prompts from the Public Pool of Prompts (P3) dataset \citep{sanh2021multitask}.
Prompts from the datasets \textsc{CommonsenseQA} \citep{talmor-etal-2019-commonsenseqa} and \textsc{CommonGen} \citep{lin-etal-2020-commongen} serve to assess the models' capability of reasoning and coherent narratives generation given some context.
We also include \textsc{AdversarialQA}, which consists of complex prompts specifically designed to mislead the models.
We randomly sample 100 examples from each dataset to create our target pool of prompts. In the case of open-domain dialog corpora, we process the dialogue into a sequence of utterances and randomly sample a turn index.
The prompt is then constructed up to that turn, and the model's task is to generate the next turn(s).

\subsection{Models}

\begin{table*}[t]
\centering
\tabcolsep=0.11cm
\caption{Details of the models used for evaluation in this work.}
\begin{tabular}{llllll}
\toprule
Model & Base model & Architecture & Size & Finetuning data & \\
\midrule
Flan-t5 & T5 & encoder-decoder &  [3B, 11B] & formal instruct & \\
Dolly-v2  & pythia & decoder-only &  [7B, 12B] &  colloquial instruct & \\
Falcon-instruct &  falcon & decoder-only &  [7B] & instruct/chat & \\
MPT-instruct & mpt & decoder-only &  [7B] & colloquial instruct/preference & \\
\bottomrule
\end{tabular}
\label{table:models}
\end{table*}

We assemble an evaluation pool comprised of several state-of-the-art and widely used language models including Flan-T5 \citep{https://doi.org/10.48550/arxiv.2210.11416} and Dolly-V2 \citep{DatabricksBlog2023DollyV2} families, as well as MPT-7b-instruct \citep{MosaicML2023Introducing} and falcon-7b-instruct \citep{refinedweb}. 
Each of these models, ranging from 3 billion to 12 billion parameters, has been selected for their demonstrated capabilities across various NLP tasks and the diversity they bring to our pool regarding architectural differences and training methodologies. 
While Flan-T5 is an encoder-decoder model, all other models are decoder only. The base models also differ between models, ranging from Pythia \citep{biderman2023pythia} to T5 \citep{chung2022scaling}. 
Table~\ref{table:models} provides additional information about each of the models.

\subsection{Completion generation} To ensure fairness in comparing models, it is essential to generate high-quality completions for each model in our evaluation pool. 
We briefly describe below the parameters we used during inference time for each model, chosen in accordance with the guidelines specified by the respective model repositories or HuggingFace model card.

For all our models, we generate completions in a zero-shot manner. 
For \textit{flan-t5-xxl} and \textit{flan-t5-xl} models, inference was run exclusively on a CPU device using the JAX framework~\citep{jax2018github}, configured for deterministic outputs via $\tau = 0$. 
In contrast, the remaining models leveraged the PyTorch framework ~\citep{NEURIPS2019_9015} for inference. 
For the dolly family, we used a temperature of $0.7$, while for MPT-7b-instruct and falcon-7b-instruct models it was set to $1$.

Due to the absence of a specific padding token for these models, we used the "\textsc{<eos>}" token as a substitute.
Completions were generated in variable batches, ranging from 8 to 50, depending on the model sizes, on an Nvidia A100 GPU with 40GB memory, to maximize computational efficiency. 
To reduce memory usage, we run inference in a \textsc{Bfloat16} setting ~\citep{kalamkar2019study}.

It is important to highlight that due to the differences in training and fine-tuning data among models and the diverse nature of our evaluation tasks, certain models faced challenges when dealing with specific datasets.
In particular \textit{MPT-7b-instruct} and \textit{falcon-7b-instruct} struggled with generating completions for prompts from the \textsc{AdversarialQA}.
To address this, we appended the phrase "Answer:" to the end of these prompts, enabling models to generate reasonable completions.
However, the quality of completions from \textit{MPT-7b-instruct} on both \textsc{AdversarialQA} and \textsc{Soda} remained subpar, necessitating the omission of these datasets in experiments involving these models.

\subsection{Human annotation collection}

\textbf{Annotator Task} We incorporate an outcome-level relative assessment (ORA) for human evaluations. This comparative method, as described by \citep{ethayarajh-jurafsky-2022-authenticity}, operates on assumptions about the rationality of human judgment, commonly employed in LLM evaluations \citep{bai2022training}: Generate pairs with outputs from two systems, have annotators select their preferred choice from each pair, and calculate a preference-based score to evaluate system performance.

In this work, we create random pairs containing completions from models $\mathsf{A}$ and $\mathsf{B}$ and ask the annotators to select which model's output they prefer or whether there is a tie between the models.
Our annotation guidelines follow a set of robust evaluation criteria focusing on task fulfillment, grammatical correctness, semantic coherence, and the naturalness of responses, as detailed in Appendix~\ref{sec:eval-guide}.
The completions generated from a predetermined set of prompts are randomly assigned as either option $\mathsf{A}$ or option $\mathsf{B}$ in the interface. This random assignment prevents any potential bias that might arise if annotators consistently favor one position over the other.  

\textbf{Random Baseline} The procedure we follow to establish a baseline for comparison between two models involves presenting prompts $P = \{p_i\}_{i=1}^N$ to annotators in a random order and subsequently collecting annotations.
To establish a reliable baseline, we take the average across $N_{perms}$ unique sequence permutations.
This technique accounts for the inherent variability when different sequences yield varying tie outcomes for the top k\% data selections.
We compare the tie rates of random sequences, averaged over $N_{perms}$, against rankings based on both KL divergence (KL) and Cross-Entropy (CE) at varying top k percentiles.
An effective metric that aptly measures the dissimilarity between two completions should manifest as reduced tie rates, particularly in the earliest, highest-ranked annotations.

\textbf{Aggregation of Preferences} 
For each unique ``A vs. B'' model comparison experiment, we collect $N_{evals}$ pairwise human preference data.
Each experiment benefits from the input of up to 10 annotators, yielding binary ranking outcomes.  
Given the subjective nature of the task and the diverse perspectives of the annotators, we observe some intra-annotator disagreement in our experiments (on average $70\%$ agreements between annotators).
To accommodate for this variability, we implement a soft voting aggregation mechanism ~\citep{davani-etal-2022-dealing}. Soft voting, often used in ensemble machine learning models, works by averaging the class probabilities of the individual annotators instead of relying on the majority count. This approach is beneficial in scenarios where there is high variability in annotator responses, as it captures a more nuanced view of the collective decision-making process.
We consider cases where there is an equal vote count for both models or when the annotators select the ``both good'' or ``both bad'' option as tie outcomes.
With soft voting, we incorporate a threshold on the averaged scores to define a tie, adding a degree of flexibility to account for close outcomes between the two models under consideration.
Examining the distribution of disagreements for each experiment, we chose a threshold of $0.2$ for comparing models within the same family and $0.1$ when comparing models from different families.

%% file: results.tex
\section{Results and Analysis}
\label{sec:exp}

An efficient ranking will aim to minimize tie outcomes among the highest-ranked instances. By prioritizing instances with a lower likelihood of inducing a tie, we can obtain early signals to determine the superior model. 
Ideally, by employing a robust ranking system, one can draw the same conclusions about model performance using a specific subset of prompts as they would have by annotating the entire dataset. 
This is particularly valuable since human annotations can be costly, and budget constraints may make it unfeasible to manually annotate all instances within a given prompt pool. Our goal is to arrive at a ranking where annotating only a subset of the top k\% of evaluation instances, where $k < N_\text{prompts}$, is sufficient to have reliable results. 

We consider two settings for evaluating the effectiveness of our proposed metrics in reducing tie outcomes: \textbf{1)} within the same model family (intra-family), and \textbf{2)} across \emph{different} model families (inter-family).
In each of these settings, we conducted two distinct experiments.
For a comprehensive view of both intra-family and inter-family experiments, including detailed win rates and tie outcomes, we refer to Appendix \ref{sec:app_rates}.

Intra-family comparisons refer to models that share analogous architectural foundations and are governed by comparable training paradigms, such as varying sizes of the same model.
In our observations, such alignment typically results in their generated log probabilities being confined to similar numerical ranges, facilitating a straightforward comparative analysis without necessitating normalization techniques.
On the other hand, in the inter-family comparisons, we observe significant variations in the range of log probabilities across models, given the same evaluation prompts set.
These disparities can be attributed to divergent training sources, architectural intricacies, or underlying methodologies. 
A normalization is imperative to ensure an objective and unbiased comparison between models across diverse sources.

\subsection{Comparing Models Within the Same Family}\label{subsec:same-family}

\begin{figure*}[t]
    \centering
    \begin{tabular}{cc}
        \subcaptionbox{\textit{flan-t5-xxl} vs. \textit{flan-t5-xl}\label{ff}}{\includegraphics[width=2.9in]{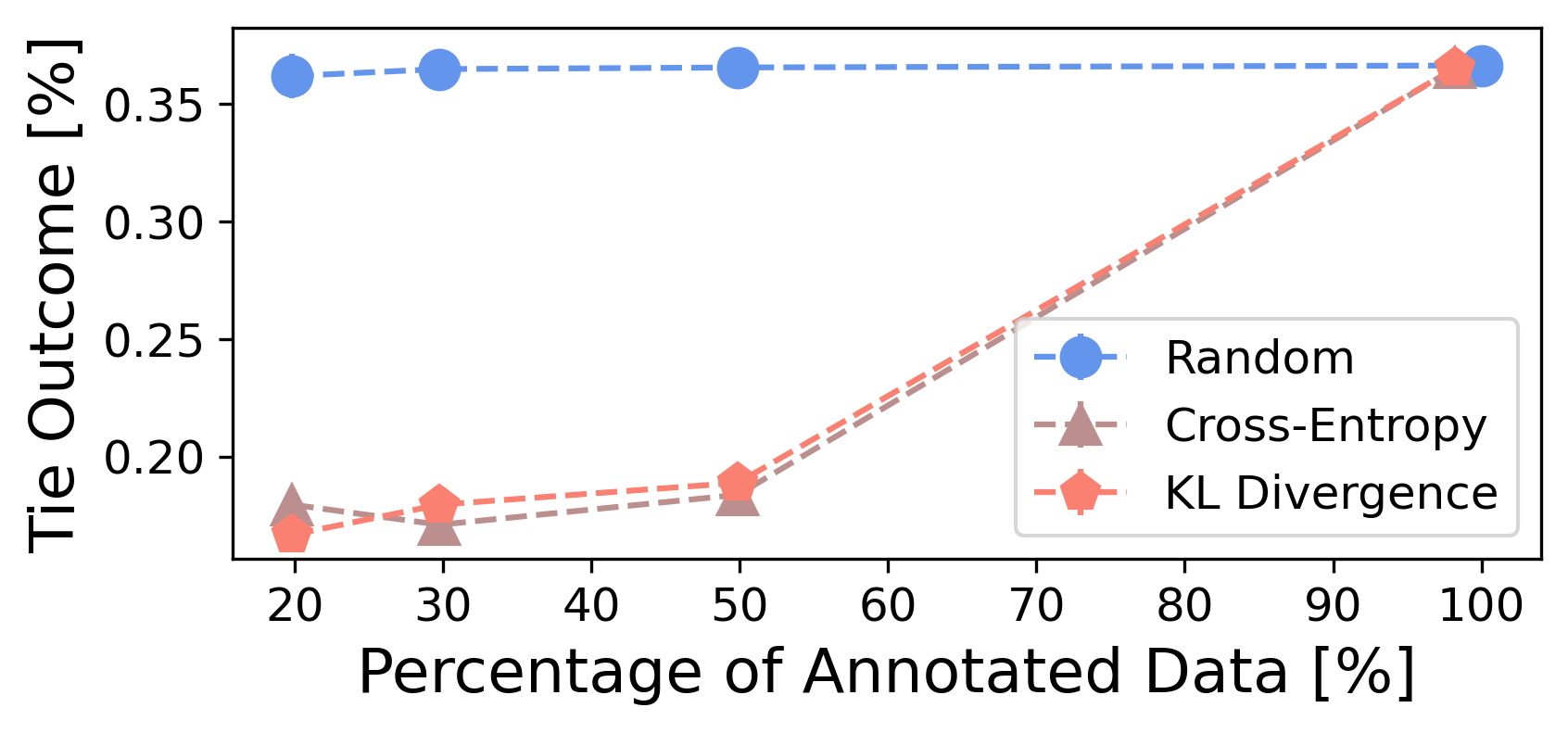}} &
        \subcaptionbox{\textit{dolly-v2-12b} vs. \textit{dolly-v2-7b} \label{dd}}{\includegraphics[width=2.9in]{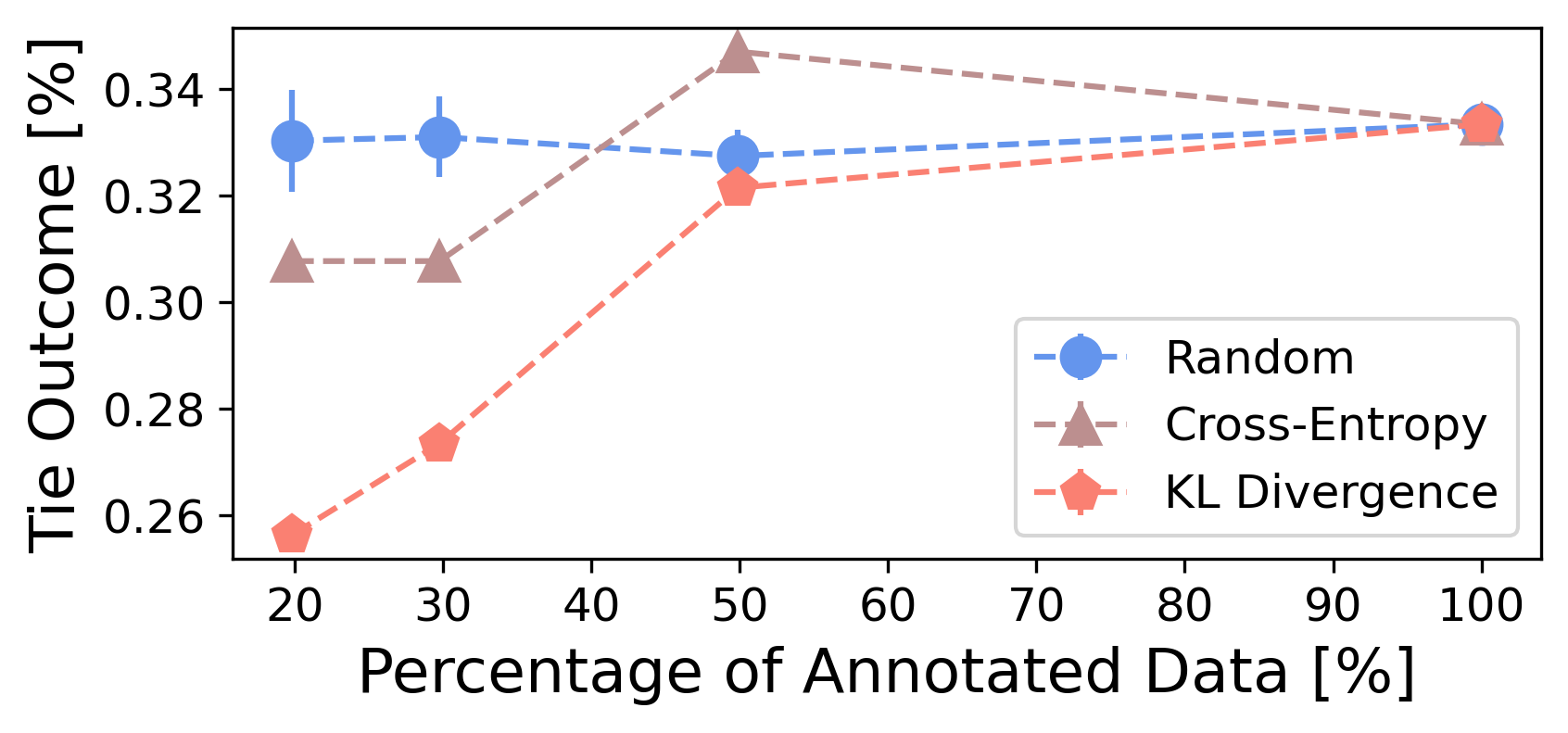}} \\
    \end{tabular}
    \caption{\textbf{Setting 1: Ranking Models from the Same Families} -- Comparison of tie ratios between KL Divergence and Cross-Entropy metrics against the average outcome of $100$ samples of randomly ordered sequences for two distinct models at various data selection percentages. The error bars represent the 95\% confidence intervals for the `Random' baseline.}
    \label{fig:intra_model_tie_rates}
\end{figure*}

We present tie rates for two sets of model comparisons: \textit{flan-t5-xxl} vs. \textit{flan-t5-xl} and \textit{dolly-v2-12b} vs. \textit{dolly-v2-7b}, across different top k percentages of collected human annotations in Figure~\ref{fig:intra_model_tie_rates}.
In our baseline, which represents a random ordering, tie rates remain consistent, showing only minor deviations around the results for 100\% annotated data.
This consistency emerges from our approach of averaging across $N_{perms}$ reordered sequences.
To demonstrate the strengths of our methodology, we set the KL Divergence and Cross-Entropy ranking metrics against our baseline.
Both dissimilarity metrics, represented by $\hat{P}_{\text{KL}}$ and $\hat{P}_{\text{CE}}$ sequences, significantly outperform the random ordering, especially between the top-20 and top-30 percentiles.

KL Divergence successfully reduces tie outcomes in the first half of the ranked set for both flan and dolly families. 
We observe reduction in ties in the top-20 percentile for models compared within the dolly family with ties reducing $6.83\%$ and $22.36\%$ relative to random ranking for Cross-Entropy and KL Divergence respectively.
As for the flan family, ties are reduced by 51.15\% in case of $\hat{P}_{\text{CE}}$ and 54.64\% for $\hat{P}_{\text{KL}}$, indicating a consistent metric performance across distinct model families.
In both experiments, we observe a significant difference between the metric-based and random rankings in the first half of the prompt list. These significant results confirm that a greater dissimilarity between completions, as indicated by higher values, corresponds to a reduced rate of ties at the top of the ranking, thereby facilitating more decisive preferences from the annotators.

\begin{table}[t]
    \centering
    \footnotesize
    \setlength\tabcolsep{5pt} 
    \caption{Percentage of decrease in number of ties for each metric relative to a random selection of prompts for comparison ``\textit{flan-t5-xxl} vs. \textit{flan-t5-xl}''.}
    \begin{tabular}{lccccc}
    \toprule
    \multicolumn{6}{c}{\textbf{\% Decrease in Ties Relative to Random Selection}}\\
    \cmidrule(lr){1-6}
    \multirow{2}{*}{\textbf{Scoring Metric}}& \multicolumn{5}{c}{\textbf{Percentile Ranking}} \\
    \cmidrule(lr){2-6}
    & \textbf{5\%} & \textbf{10\%} & \textbf{20\%} & \textbf{30\%} & \textbf{50\%} \\
    \midrule
    KL Divergence  & 57.81 & 51.12 & 54.64 & 51.16 & 48.72 \\
    Cross-Entropy & 29.68 & 30.17 & 51.15 & 53.49 & 50.10 \\
    \bottomrule
    \end{tabular}
    \label{tab:flan-improv}
\end{table}

\subsection{Comparing Models from Different Families}
\label{subsec:diff-family}

A natural question that arises following our analysis of model specific instance prioritization is whether this analysis is transferable to comparing models from two different model settings. This setting is similar to the one described in Section \ref{subsec:same-family} but involves ranking models from different families.

To compare models across different families, we first normalize probability ranges by adopting a min-max scaling approach as expressed in Equation \ref{eq:minmax}. $P_{\text{min}}$ and $P_{\text{max}}$ represent the minimum and maximum probabilities across both models in comparison, respectively.
\begin{equation}
    \label{eq:minmax}
    P_{\text{norm}} = \frac{P - P_{\text{min}}}{P_{\text{max}} - P_{\text{min}}}
\end{equation}

This standardization method enables the computation of our metrics on a comparable scale across different model families. Once these adjustments are implemented, we follow the same ranking and comparison procedures as in the inter-family case. 

\begin{figure}[t]
    \centering
    \begin{tabular}{cc}
        \subcaptionbox{\textit{flan-t5-xxl} vs. \textit{dolly-v2-12b}\label{fd}}{\includegraphics[width=2.9in]{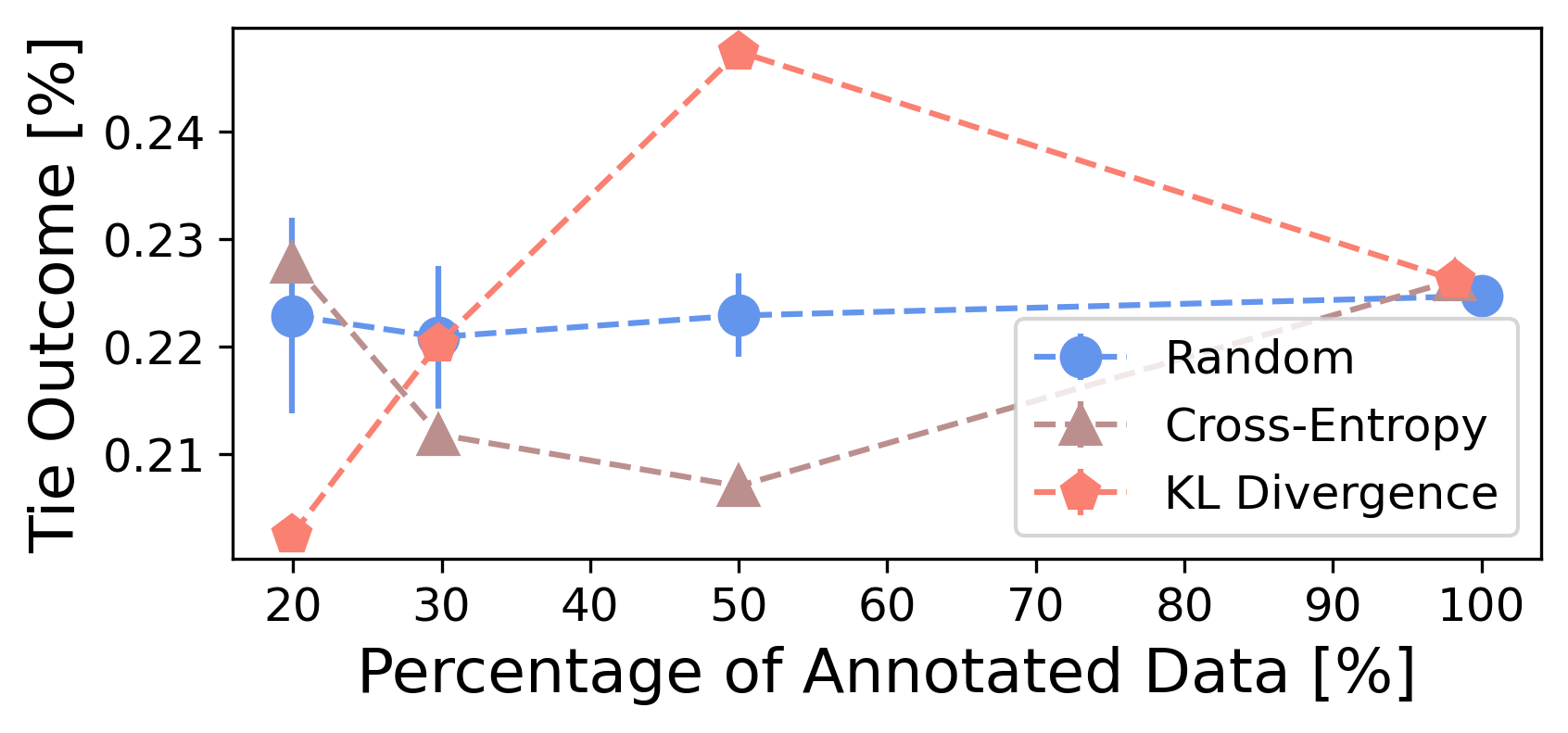}} &
        \subcaptionbox{\textit{MPT-7b} vs. \textit{falcon-7b}\label{mf}}{\includegraphics[width=2.9in]{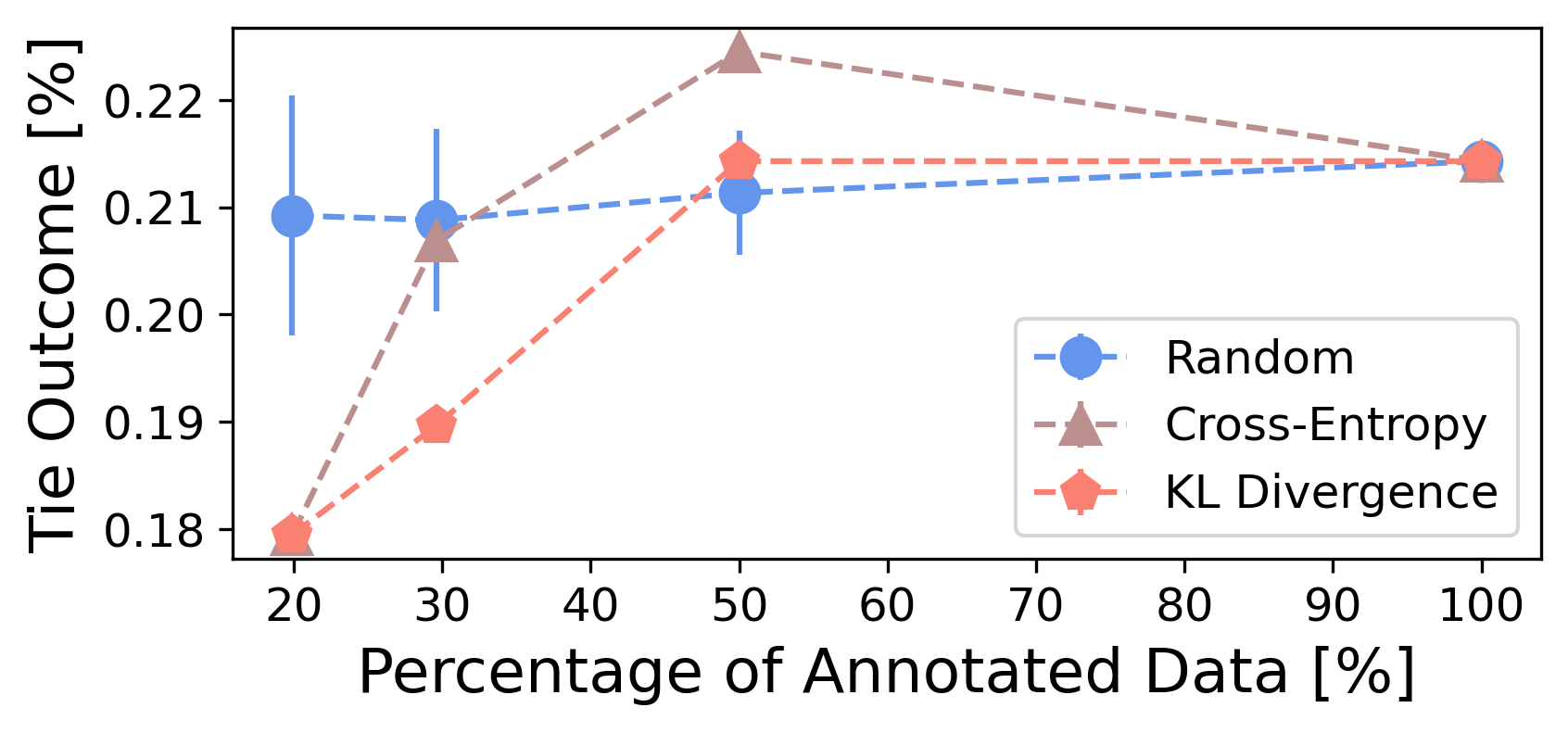}}
    \end{tabular}
    \caption{\textbf{Setting 2: Ranking Models from Different Families} -- Comparison of tie ratios between KL Divergence and Cross-Entropy metrics against the average outcome of $100$ samples of randomly ordered sequences for two distinct models at various data selection percentages. The error bars represent the 95\% confidence intervals for the `Random' baseline.}
    \label{fig:inter_model_tie_rates}
\end{figure}

Figure~\ref{fig:inter_model_tie_rates} illustrates our intra-family main results.
for the model comparison between ``\textit{flan-t5-xxl} and \textit{dolly-v2-12b}'' at 20\% annotated data, the Cross-Entropy metric reduces the tie rates by $9.85\%$. 
The KL Divergence metric exhibits performance similar to random ranking, with a marginal increase in the tie rate of $1.4\%$.
At 30\%, both metrics conform to the anticipated trend, with Cross-Entropy marginally outperforming KL Divergence.

In the comparison between ``\textit{MPT-7b} and \textit{falcon-7b}'', the most significant reduction in tie outcomes is observed at 20\% annotated data, followed by a gradual increase up to 50\%. Here, KL outperforms CE, in agreement with the results of the inter-family experiments presented in section \ref{subsec:same-family}.
It is imperative to note that while the improvements in this setting might not be as pronounced as in the same family models, it is primarily due to the inherently lower tie rates in the different family comparisons. Nonetheless, even in this challenging scenario, our metrics evidently provide a noticeable advantage when examining the top percentile.

\subsection{Comparing the two settings}
We observe more efficiency gains in {setting 1} where we compare models from the same family as opposed to intra-family comparisons in {setting 2}. 
Typically models in \textbf{setting 1} tend to exhibit similar generalization behaviors due to similar optimization, finetuning data, and base models, and it is challenging to distinguish improvements between them. 
Here, we observe the most significant gains in efficiency by leveraging our method. 

However, when comparing models from different families (\textbf{setting 2}), the generated outputs can vary significantly based on specific tasks or prompts \citep{strobelt-etal-2021-lmdiff}.
When models are from different model families, there is typically a more pronounced performance spread which make it easier for human annotators to distinguish between model outputs. 
Therefore, there are fewer tie outcomes, leading to fewer expected benefits from prioritization.
Our findings indicate that our method can still bring meaningful efficiency improvements even in this challenging setting.
It is crucial to underscore the real-world implications of these findings.
In practical scenarios, especially within production environments, it is more common to study comparisons within model families. The primary objective in such instances is to rank and evaluate various iterations of a single model, aiming to select the most optimal version for deployment.
In such cases, using our proposed metrics to rank model completions substantially diminishes tie occurrences in top comparisons, resulting in cost savings in terms of annotation.

\subsection{Elo Score Robustness}

\begin{figure}[!t]
    \centering
    \begin{subfigure}[t]{0.85\textwidth}
        \includegraphics[width=\textwidth]{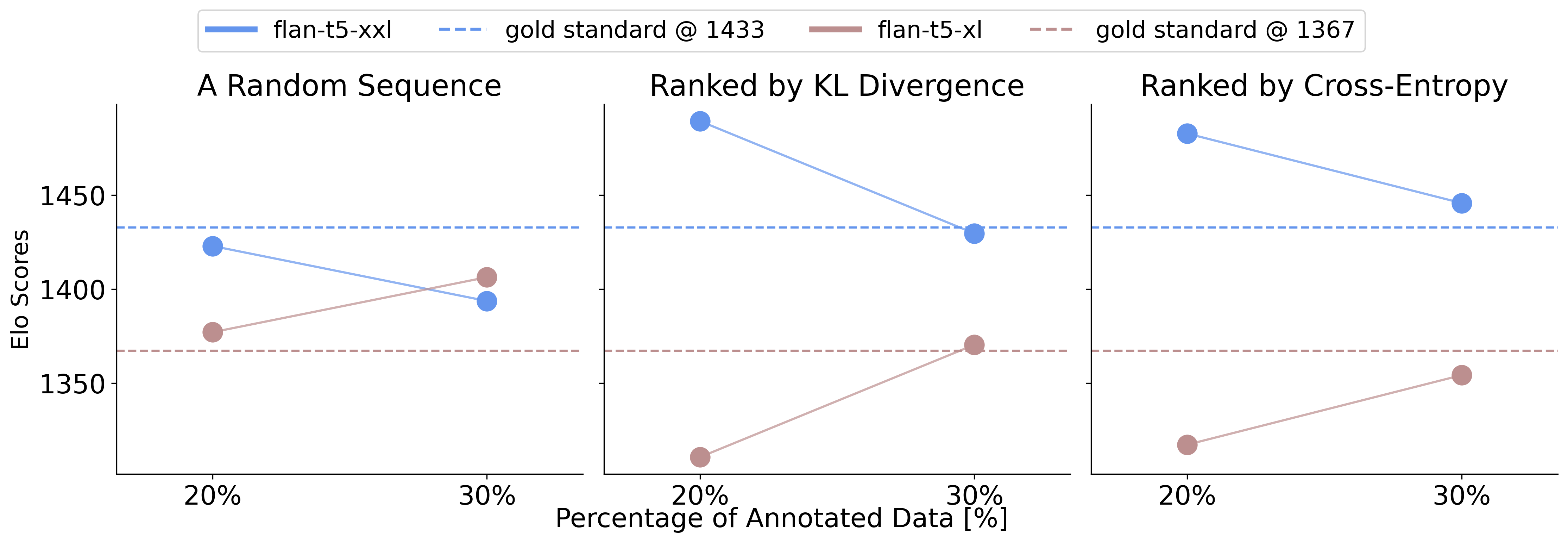}
        \caption{\textit{flan-t5-xxl} vs. \textit{flan-t5-xl}}
        \label{fig:flan-elo-scores}
    \end{subfigure}

    \vspace{1em}

    \begin{subfigure}[t]{0.85\textwidth}
        \includegraphics[width=\textwidth]{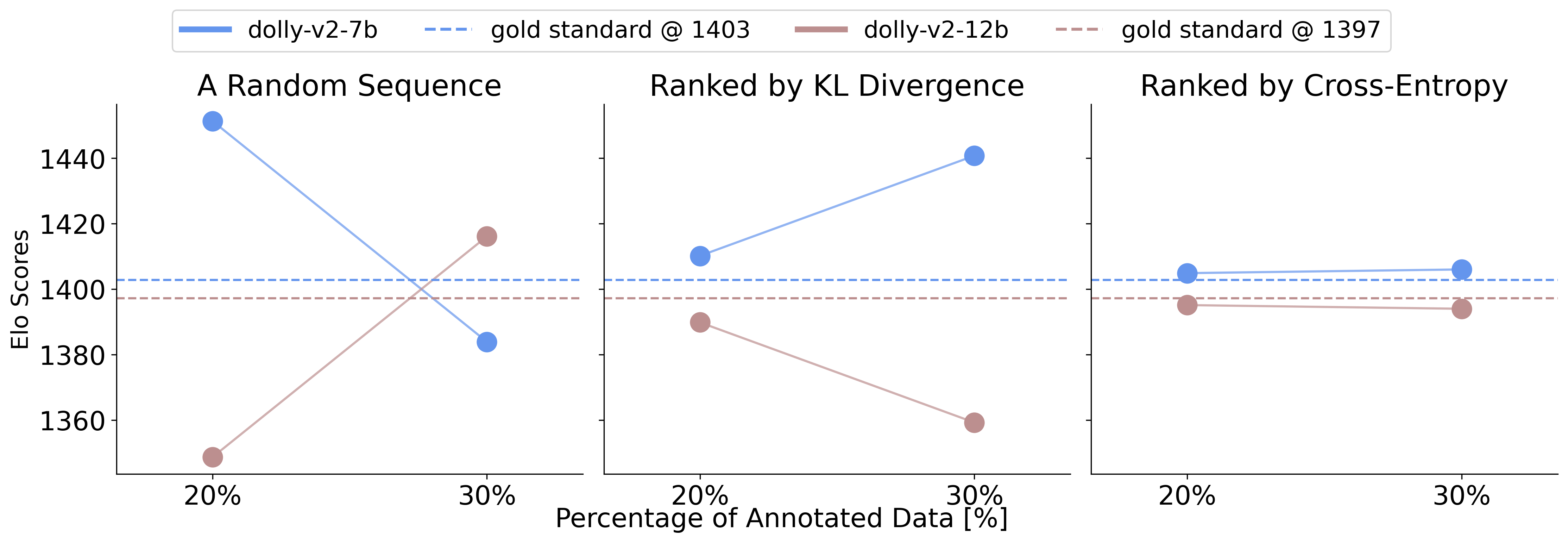}
        \caption{\textit{dolly-v2-7b} vs. \textit{dolly-v2-12b}}
        \label{fig:dolly-elo-scores}
    \end{subfigure}
    \caption{Elo scores of intra-family comparisons using the top 20\% and top 30\% subsets of evaluation data. Dashed lines represent the Elo scores, determined using 100\% of human annotation data, establishing our `Gold Standard' models ranking. For further details, we refer to Table \ref{tab:elo_scores}.}
    \label{fig:flans-elo-scores-percentages}
\end{figure}

In this section, we investigate how our findings impact the robustness of Elo rating system, previously introduced in Section \ref{sec:elo}.
To update the Elo ratings of two LLMs, the models engage in a head-to-head match for each prompt and annotators express their preference of the model completions.
Tie outcomes can occur due to individual annotator indecision as well as inter-annotator disagreement.
Selecting one model's output as superior in a pairwise comparison provides a robust signal for the Elo rating to adjust accordingly.
However, in the case of a tie, the adjustment in rating points is typically smaller, as ties are considered less informative about the relative strengths of the models.
As a result, both models end up gaining or losing the same number of rating points.
A tie treats both models' quality as equally uncertain, which does not provide valuable information for ranking the models.
The order of prompts affects how quickly and accurately the ratings converge to reflect the true skill levels of the models.

To evaluate the impact of our proposed prompt order on Elo scores, we begin by establishing a consistent baseline. We achieve this by averaging the Elo scores over 100 permutations for each of our four experiments, comparing intra- and inter-family models. These averages serve as our `Gold Standard' model rankings, as summarized in Table \ref{tab:elo_scores}.

We next evaluate the efficacy of our ranking metrics by calculating Elo scores over prompts ordered by CE and KL metrics.

In Figure~\ref{fig:flans-elo-scores-percentages}, we present Elo scores, focusing on the top percentile of the evaluation set. 
When using randomly sampled prompts, the Elo scores display unpredictable behavior, switching model rankings frequently in the initial stages.
For instance, we can observe that with a random sequence, relying on the Elo ratings with only 30\% of the data is insufficient to assess the relative performance of the two models:
the random sequence Elo scores suggest that \textit{flan-t5-xl} beats \textit{flan-t5-xxl} (Figure~\ref{fig:flans-elo-scores-percentages} (a)) and \textit{dolly-v2-12b} beats \textit{dolly-v2-7b} (Figure~\ref{fig:flans-elo-scores-percentages} (b)). 
However the dashed lines in the figures indicates that the correct ranking is the opposite for both cases.
This is mainly due to the appearance of prompts with tie outcomes in the top percentile.
In contrast, our metrics demonstrate a more reliable and consistent performance. 
Both KL Divergence and Cross-Entropy effectively rank the two models from the early stages of evaluation and maintain this consistency throughout the entire process.
As a result, we can confidently employ a smaller subset of the evaluation set sorted by our metrics, and achieve Elo scores equivalent to the Gold Standard obtained using the full dataset. 
This is shown to be not the case when using a random sequence of prompts.
Our proposed evaluation provides a fast and thorough strategy to assess models performance relative to each other, enhancing trust in results even when working with smaller datasets.

\begin{table}[t]
    \centering
    \sisetup{separate-uncertainty=true}
    \begin{tabular}{lS[table-format=6.4(2)]}
    \toprule
    \textbf{Experiment} & {\textbf{Elo Score $S$ }}\\
    \midrule
    Flan-t5-xxl & 1432.69 \pm 2.34 \\
    Flan-t5-xl & 1367.31 \pm 2.34 \\
    \midrule
    Dolly-v2-7b & 1402.82 \pm 2.06 \\
    Dolly-v2-12b & 1397.18 \pm 2.06 \\
    \midrule
    Flan-t5-xxl & 1481.70 \pm 2.06 \\
    Dolly-v2-12b & 1318.30 \pm 2.06 \\
    \midrule
    Falcon-7b-instruct & 1513.45 \pm 1.85 \\
    MPT-7b-instruct & 1286.55 \pm 1.85 \\
    \bottomrule
    \end{tabular}
    \caption{ `Gold Standard' Elo scores, accompanied by their respective Standard Errors of the Mean (SEM), derived from our human evaluation experiments. These scores are computed from an average over $N_\text{perms} = 100$ permutations.}
    \label{tab:elo_scores}
\end{table}

\subsection{Limitations}

Our methodology focuses on prioritizing evaluation instances showcasing distinct model behaviors, with the goal of minimizing tie outcomes.
While this strategy is aimed at optimizing the evaluation process, especially when resources are limited, it could inherently favor certain data points, possibly leading to some biases.
By prioritizing pronounced differences, we risk over-representing certain challenges and under-representing areas where models have consistent outputs.
In particular, prompts that help to establish the lower and upper performance bounds of models may be underrepresented in our ranking methodology.
It's important to note that our proposed methodology is designed to prioritize annotation within the constraints of one's budget, rather than using it to make decisions about sample exclusion.

%% file: related.tex
\section{Related work}
\label{sec:rel}

\subsection{Evaluation of LLMs} Automatic evaluation methods of NLP systems has been a longstanding goal for researchers and practitioners  \citep{lan2020pone,pang2020towards,deriu2020spot}. Recently, using language models as Natural Language Generation (NLG) evaluation metric, such as embedding-based metrics \citep{zhang2019bertscore,zhao2019moverscore} that evaluate the semantic similarity via pre-trained BERT model, LLM-based metrics \citep{yuan2021bartscore,fu2023gptscore} that evaluate the quality of the NLG model outputs, learning-metric~\citep{chen-etal-2022-storyer} that mimics human preference when judging a story, receives increasing attention since it offers a proxy for human-related judgement. \citet{lin2023llmeval} introduced a method called LLM-Eval that assesses multiple dimensions of conversation quality, including content, grammar and relevance, without the need for human references or prompts.

However, these metrics still struggle to fully encode the quality aspects of generated text \citep{chaganty2018price,celikyilmaz2020evaluation}. Furthermore, large language modelling has led to models which are widely deployed and often used for many different tasks. This has posed challenges for traditional evaluation. Instead of only evaluating for a single task, models are often expected to perform well at many tasks. This has prompted a shift towards relying on automated benchmark approaches that collate many tasks measuring different narrow properties \citep{gehrmann2022gemv2,wang2022adversarial}.
Recent benchmarks \citep{liang2022holistic,zheng2023judging,gu2023xiezhi} are increasingly supplementing automated metrics with human judgement.
Compared to automatic evaluation, human evaluation closely resembles real-life scenarios and offers more thorough and precise feedback \citep{bubeck2023sparks,ziems2023can,bang2023multitask}.

\subsection{Improving quality of human-in-the-loop evaluation} 
The importance of analyzing and standardizing human evaluation methods in text generation tasks has been gaining more attention due to a lack of consensus on how to qualitatively evaluate NLG systems \citep{van2019best,knowles-2021-stability} and the renewed importance of annotations in preference learning and alignment optimization.
Previously, various annotation frameworks have been developed to make the annotation process robust and reusable \citep{ide-etal-2003-international,DBLP:journals/tal/ChiarcosDGLLRS08,sabou-etal-2014-corpus}. 
Annotating datasets requires attention to various aspects of the problem, including nuances of the language \citep{bergman-diab-2022-towards}, user adaptivity \citep{dipper-etal-2004-towards}, annotator bias \citep{thorn-jakobsen-etal-2022-sensitivity}, and domain knowledge \citep{yada-etal-2020-towards}. 
While our primary focus is to minimize human-in-the-loop feedback to effectively distinguish between models, our prompt prioritization approach also ensures robust performance evaluation.

Different task designs and data collection methods impact the consistency of collected judgments \cite{novikova2018rankme,santhanam2019towards}. Comparative approaches have proved successful in contrast to direct evaluation \citep{novikova2018rankme}, but tend to require multiple head-to-head comparisons to achieve statistical significance \citep{celikyilmaz2020evaluation}.
Initiatives like GENIE~\citep{khashabi2021genie} strives to automate and standardize the human evaluation of various NLG systems, with the goal of creating a human evaluation leaderboard to track LLM performance. Another approach has introduced by \citet{gehrmann2021gem}, uses GENIE's infrastructure for conducting human evaluations while adopting its own human evaluation techniques. \cite{peyrard-etal-2021-better} focuses on aggregation methods of human feedback in pairwise comparisons of NLG systems. While these works concentrate on the format of annotation that leads to the highest quality, our focus is instead to provide a general and scalable framework to prioritize instances for feedback in a manner that is agnostic to the annotation tool.

\subsection{Improving efficiency of human-in-the-loop evaluation} 
Collecting human judgments remains a core component of evaluation of NLP systems. However, this process can be both expensive and time-consuming. Efforts have been made to address this issue and devise more efficient evaluation strategies.
\citet{thorleiksdottir2021dynamic} conducted a comprehensive analysis of common labelling strategies, considering varying difficulties for model comparisons. Their analysis aiming to commence the design of new evaluation methods that require a reduced number of annotations. 
Another approach to improving efficiency in evaluation involves employing Item Response Theory (IRT) for assessing NLG systems. \citet{sedoc2020item} employed IRT by collecting binary comparisons of system responses and reduced the overall number of samples required for model assessment.
In contrast, the focus of our work is on optimizing the annotation effort when comparing two models, ensuring a confident model decision by ranking prompts and completion pairs to be used in human evaluation. By strategically selecting and prioritizing the most informative samples for evaluation, our proposed approach aims to make the evaluation process more efficient and resource-effective.

\subsection{Prioritizing data instances} 
Much of NLP development has involved making heuristic choices about which data to prioritize for training and optimization.
The selection processes to determine what goes into large scale datasets have centered on rule based filters and heuristics \citep{bane-etal-2022-comparison}, such as removing Reddit threads with fewer than three responses \citep{zhang2022opt}, removing entire components of the PILE based upon perceived quality or blocklists \citep{dodge2021documenting} or filtering out documents that fall outside specific word count or vocabulary criteria \citep{rae2022scaling}. 
In contrast to these heuristics, our focus is on scalable and rigorous techniques to prioritize a subset of prompt and completions for human annotation during the evaluation stage. 
While recent research has concentrated on the prioritizing datasets during the training and fine-tuning stages, typically through the application of quality filters as selection criteria, our emphasis is on the prioritization of evaluation instances.
For instance, models like Alpaca \citep{taori2023alpaca}, Vicuna \citep{chiang2023vicuna}, and Koala \citep{vu2023koala} all have as a base the LLaMA model \citep{touvron2023llama} combined with instruction data generated from existing large language models.

%% file: appendix.tex
\section{Human Feedback Collection}
\label{sec:app_human_feedback}
\subsection{Evaluations}
\label{sec:eval-samples}

This subsection showcases random evaluation instances to highlight performance disparities between models within the same family: \textit{flan-t5-xxl} vs. \textit{flan-t5-xl}, and \textit{dolly-v2-7b} vs. \textit{dolly-v2-12b}, across diverse datasets.
Drawing samples from both the top and bottom 20\% according to our ranking metrics, we illustrate the capability of these metrics to systematically arrange evaluations based on the dissimilarity of completions in ``A vs. B'' comparisons.
Conversely, examples from the bottom percentile display instances where the models generate outputs of similar quality, increasing the likelihood of tied outcomes.
Examples from the top 20\% highlight dissimilar completions, demonstrating the discriminative power of our ranking metrics.

\begin{table}
    \centering
    \footnotesize
    \caption{Comparison of \textit{flan-t5-xxl} vs. \textit{flan-t5-xl}}
    \begin{tabularx}{\textwidth}{|l|X|}
        \hline
        \rowcolor{gray!30}
        \multicolumn{2}{|l|}{Ranked by KL Divergence} \\
        \hline
        \hline
        \multicolumn{2}{|c|}{\textbf{\textsc{Soda} - Top 20\%}} \\
        \hline
        \textbf{Prompt} & Rosalina went to the ticket counter at the tennis stadium and bought two tickets to the match that was going to start in an hour. She was excited to see her favorite player, Rafael Nadal, compete against his longtime rival, Roger Federer.\\
        & Rosalina: Hi, I would like to buy two tickets to the match that's happening in an hour.\\
        & Ticket Seller: That will be $180$.\\
        & Rosalina: Okay, here you go.\\
        & Ticket Seller: Thank you. Enjoy the match!\\
        & Rosalina: Thanks! I'm really excited to see it. Nadal is my favorite player.\\
        & Ticket Seller: \\
        \hline
        \textbf{Completion A} & Ticket Seller: Have a good time!\\
        \hline
        \textbf{Completion B} & Rosalina: I'm sure he'll win.\\
        \hline
        \multicolumn{2}{|c|}{\textbf{\textsc{Soda} - Bottom 20\%}} \\
        \hline
        \textbf{Prompt} & Hashim never refuses when someone asks her for a favor because she wants to be seen as a good person. She knows that by saying "yes" to everything, people will appreciate her more and think of her as being helpful. Hashim also feels that it is her duty to help others, even if it means going out of her way or inconveniencing herself.\\
        & Hashim: Hey, how are you doing?\\
        & Friend: I'm good, thanks. How about you?\\
        & Hashim: I'm good. Hey, do you need any help with anything?\\
        & Friend: No, I don't think so. Thanks for offering though.\\
        & Hashim: No problem. I just wanted to help out if I could.\\
        & Friend: That's really nice of you.\\
        & Hashim: Yeah, I just feel like it's the right thing to do. Helping others is important to me.\\
        & Friend: I can see that. It's one of the things I like about you. You're always willing to help out, no matter what.\\
        & Hashim: Thanks. I just think that it's important to be there for people when they need help. We all have to stick together and help each other out.\\
        & Friend: \\
        \hline
        \textbf{Completion A} & Yeah, I agree.\\
        \hline
        \textbf{Completion B} & That's a great philosophy.\\
        \hline
    \end{tabularx}
\end{table}

\begin{table}
    \centering
    \footnotesize
    \caption{Comparison of \textit{flan-t5-xxl} vs. \textit{flan-t5-xl}}
    \begin{tabularx}{\textwidth}{|l|X|}
        \hline
        \rowcolor{gray!30}
        \multicolumn{2}{|l|}{Ranked by KL Divergence} \\
        \hline
        \hline
        \multicolumn{2}{|c|}{\textbf{\textsc{CommonsenseQA} - Top 20\%}} \\
        \hline
        \textbf{Prompt} & Here's a question: Where would one find a snake in a cage? \\
        & Here are possible answers to this question: \\
        & - tropical forest \\
        & - oregon \\
        & - woods \\
        & - pet store \\
        & - louisiana \\
        & I believe the correct choice is "pet store", here's why: \\
        \hline
        \textbf{Completion A} & pet stores sell animals \\
        \hline
        \textbf{Completion B} & pet store would one find a snake in a cage \\
        \hline
        \multicolumn{2}{|c|}{\textbf{\textsc{CommonsenseQA} - Bottom 20\%}} \\
        \hline
        \textbf{Prompt} & Here's a question: What uses a ribbon to put words on paper? \\
        & Here are possible answers to this question: \\
        & - wrapping paper \\
        & - girl's hair \\
        & - bath \\
        & - floral arrangement \\
        & - typewriter \\
        & I believe the correct choice is "typewriter", here's why: \\
        \hline
        \textbf{Completion A} & typewriters are mechanical devices for writing \\
        \hline
        \textbf{Completion B} & typewriters use ribbons to write \\
        \hline
        \multicolumn{2}{|c|}{\textbf{\textsc{CommonGen} - Top 20\%}} \\
        \hline
        \textbf{Prompt} & Put the concepts together to form a sentence: cart, road, ride. \\
        \hline
        \textbf{Completion A} & A man riding a cart down a road. \\
        \hline
        \textbf{Completion B} & A man riding a horse and cart down a road. \\
        \hline
        \multicolumn{2}{|c|}{\textbf{\textsc{CommonGen} - Bottom 20\%}} \\
        \hline
        \textbf{Prompt} & Put the concepts together to form a sentence: sit, car, backseat. \\
        \hline
        \textbf{Completion A} & A man sits in the backseat of a car. \\
        \hline
        \textbf{Completion B} & A man sits in the backseat of a car. \\
        \hline
    \end{tabularx}
\end{table}

\begin{table}
    \centering
    \footnotesize
    \caption{Comparison of \textit{flan-t5-xxl} vs. \textit{flan-t5-xl}}
    \begin{tabularx}{\textwidth}{|l|X|}
        \hline
        \rowcolor{gray!30}
        \multicolumn{2}{|l|}{Ranked by KL Divergence} \\
\hline
\hline
\multicolumn{2}{|c|}{\textbf{\textsc{AdversarialQA} - Top 20\%}} \\
\hline
\textbf{Prompt} & Extract the answer to the question from the following context.\newline

Question: What broadcast a signal?\newline

Context: The city is served by the Tyne and Wear Metro, a system of suburban and underground railways covering much of Tyne and Wear. It was opened in five phases between 1980 and 1984, and was Britain's first urban light rail transit system; two extensions were opened in 1991 and 2002. It was developed from a combination of existing and newly built tracks and stations, with deep-level tunnels constructed through Newcastle city centre. A bridge was built across the Tyne, between Newcastle and Gateshead, and opened by Queen Elizabeth II in 1981. The network is operated by DB Regio on behalf of Nexus and carries over 37 million passengers a year, extending as far as Newcastle Airport, Tynemouth, South Shields and South Hylton in Sunderland. In 2004, the company Marconi designed and constructed the mobile radio system to the underground Metro system. The Metro system was the first in the UK to have mobile phone antennae installed in the tunnels.\\
\hline
\textbf{Completion A} & Marconi \\
\hline
\textbf{Completion B} & antennae \\
\hline
\multicolumn{2}{|c|}{\textbf{\textsc{AdversarialQA} - Bottom 20\%}} \\
\hline
\textbf{Prompt} & Extract the answer to the question from the following context.\newline

Question: Area shortages meant a shortage of what?\newline

Context: In 1939, c. 1,300,000 people lived in Warsaw, but in 1945 – only 420,000. During the first years after the war, the population growth was c. 6\%, so shortly the city started to suffer from the lack of flats and of areas for new houses. The first remedial measure was the Warsaw area enlargement (1951) – but the city authorities were still forced to introduce residency registration limitations: only the spouses and children of the permanent residents as well as some persons of public importance (like renowned specialists) were allowed to get the registration, hence halving the population growth in the following years. It also bolstered some kind of conviction among Poles that Varsovians thought of themselves as better only because they lived in the capital. Unfortunately this belief still lives on in Poland (although not as much as it used to be) – even though since 1990 there are no limitations to residency registration anymore. \\
\hline
\textbf{Completion A} & flats \\
\hline
\textbf{Completion B} & new houses \\
\hline
\end{tabularx}
\end{table}

\begin{table}
    \centering
    \footnotesize
    \caption{Comparison of \textit{dolly-v2-7b} vs. \textit{dolly-v2-12b}}
    \begin{tabularx}{\textwidth}{|l|X|}
        \hline
        \rowcolor{gray!30}
        \multicolumn{2}{|l|}{Ranked by Cross-Entropy} \\
\hline
\hline
\multicolumn{2}{|c|}{\textbf{\textsc{Soda} - Top 20\%}} \\
\hline
\textbf{Prompt} & Benzion joins an activity because Benzion wants to meet new people and get out of the house. Joining an activity will help Benzion connect with other people who have similar interests. \newline

Benzion: Hi, I'm Benzion. \newline
Other members: \\
\hline
\textbf{Completion A} &  Hi, I'm Joe, Slack, and Eric. Welcome to the group activity!\\
\hline
\textbf{Completion B} &  Hi, I'm Benzion. I'm looking to meet other people who are interested in the same activities as me.\\
\hline
\multicolumn{2}{|c|}{\textbf{\textsc{Soda} - Bottom 20\%}} \\
\hline
\textbf{Prompt} & Asael has a friend over. His friend is a boy named Tim. Asael and Tim are playing video games together in Asael's room.

Asael: Yeah, I just got this game yesterday and I've been dying to play it with someone.

Tim: No problem, dude. I'm always up for some video games.

Asael: Alright, cool. So, what do you think of this game so far?

Tim: It's actually really fun! I like the graphics and the gameplay is solid.

Asael: Yeah, I know! I was worried it was going to be too easy, but it's definitely challenging.

Tim: Yeah, definitely. But that's what makes it more fun in my opinion. If a game is too easy, it gets boring pretty quickly. But this one is just hard enough to keep you engaged.

Asael: Exactly! And there are so many different levels to play through too. Have you gotten very far?

Tim: Yeah, I'm on the fourth level right now.

Asael: Oh wow, you're way ahead of me! I'm still on the second level.
Tim: Well, it's not too difficult once you get the hang of it.

Asael: Yeah, I think I'm starting to get a little better. So, how do you like the new controller?

Tim: It's okay, I guess. The buttons are a little too close together for my liking though.

Asael:   \\
\hline
\textbf{Completion A} & Asael and Tim are playing video games together in Asael's room.\\
\hline
\textbf{Completion B} &  Asael has a friend over. His friend is a boy named Tim. Asael and Tim are playing video games together in Asael's room. \\
\hline
\end{tabularx}
\end{table}

\begin{table}
    \centering
    \footnotesize
    \caption{Comparison of \textit{dolly-v2-7b} vs. \textit{dolly-v2-12b}}
    \begin{tabularx}{\textwidth}{|l|X|}
        \hline
        \rowcolor{gray!30}
        \multicolumn{2}{|l|}{Ranked by Cross-Entropy} \\
\hline
\hline
\multicolumn{2}{|c|}{\textbf{\textsc{CommonsenseQA} - Top 20\%}} \\
\hline
\textbf{Prompt} & Here's a question: A mountie got off at a subway stop.  What city might he be in? \newline

Here are possible answers to this question:\newline
- urban area\newline
- metropolis\newline
- chicago\newline
- new york city\newline
- toronto\newline

I believe the correct choice is "toronto", here's why:\\
\hline
\textbf{Completion A} &  The subway stop the mountie got off at is called St. George station.  St. George is a suburb of Toronto, Ontario, Canada.\\
\hline
\textbf{Completion B} &  A mountie is a member of the national police force of Canada, so he is probably from Toronto.  Toronto is the largest city in Canada, so it is a good choice. \\
\hline
\hline
\multicolumn{2}{|c|}{\textbf{\textsc{CommonsenseQA} - Bottom 20\%}} \\
\hline
\textbf{Prompt} & Here's a question: What does a gambler do that causes him or her to be unhappy? \newline

Here are possible answers to this question:\newline
- play cards\newline
- double winnings\newline
- lose money\newline
- play poker\newline
- to win the prize\newline

I believe the correct choice is "lose money", here's why:\\
\hline
\textbf{Completion A} & A gambler who plays cards for a living might be unhappy if they consistently win money, because they have to go back to work. A gambler who plays poker might be unhappy if they lost all of their money, because they lost a game. A gambler who plays poker for a prize might be unhappy if they won the prize, because they would have to give up their prize to the prize provider. A gambler who plays cards for money might be unhappy if they win too much money, because they would have to spend their winnings and then go back to work. A gambler who plays cards for fun might be unhappy because they are not having fun.\\
\hline
\textbf{Completion B} & Playing cards, poker, and other games of chance can be very enjoyable, but they can also be very stressful.  Stressful situations are often unpleasant, and can lead to negative emotions such as anger, anxiety, and depression.  If you are a gambler and any of these things happen to you, you may be experiencing unpleasant emotions that are caused by your gambling.  If this is the case, you should seek help from a therapist or other mental health professional.  They can help you identify and work through the causes of these negative emotions.  If you continue to experience negative emotions despite working through these causes, you may want to consider changing your gambling habits.\\
\hline
\end{tabularx}
\end{table}

\begin{table}
    \centering
    \footnotesize
    \caption{Comparison of \textit{dolly-v2-7b} vs. \textit{dolly-v2-12b}}
    \begin{tabularx}{\textwidth}{|l|X|}
        \hline
        \rowcolor{gray!30}
        \multicolumn{2}{|l|}{Ranked by Cross-Entropy} \\
\hline
\hline
\multicolumn{2}{|c|}{\textbf{\textsc{AdversarialQA} - Top 20\%}} \\
\hline
\textbf{Prompt} & Extract the answer to the question from the following context.\newline

Question: What happens to those that pay too little?\newline

Context: In a purely capitalist mode of production (i.e. where professional and labor organizations cannot limit the number of workers) the workers wages will not be controlled by these organizations, or by the employer, but rather by the market. Wages work in the same way as prices for any other good. Thus, wages can be considered as a function of market price of skill. And therefore, inequality is driven by this price. Under the law of supply and demand, the price of skill is determined by a race between the demand for the skilled worker and the supply of the skilled worker. "On the other hand, markets can also concentrate wealth, pass environmental costs on to society, and abuse workers and consumers." "Markets, by themselves, even when they are stable, often lead to high levels of inequality, outcomes that are widely viewed as unfair." Employers who offer a below market wage will find that their business is chronically understaffed. Their competitors will take advantage of the situation by offering a higher wage the best of their labor. For a businessman who has the profit motive as the prime interest, it is a losing proposition to offer below or above market wages to workers.\\
\hline
\textbf{Completion A} &  In a purely capitalist mode of production, the workers wages are not controlled by the professional and labor organizations, but rather by the market. Wages are a function of market price of skill. In this context, inequality is driven by this price. Under the law of supply and demand, the price of skill is determined by a race between the demand and supply of the skilled worker. Markets can also concentrate wealth, pass environmental costs on to society, and abuse workers and consumers. By themselves, even when they are stable, often lead to high levels of inequality, outcomes that are widely viewed as unfair. Employers who offer a below market wage will find that their business is chronically understaffed. Their competitors will take advantage of the situation by offering a higher wage the best of their labor. For a businessman who has the profit motive as the prime interest, it is a losing proposition to offer below or above market wages to workers. \\
\hline
\textbf{Completion B} &  The subway stop the mountie got off at is called St. George station.  St. George is a suburb of Toronto, Ontario, Canada.\\
\hline
\hline
\multicolumn{2}{|c|}{\textbf{\textsc{AdversarialQA} - Bottom 20\%}} \\
\hline
\textbf{Prompt} & Here's a question: What could happen if someone is seeing new presents at a birthday party?\newline

Here are possible answers to this question:\newline
- envy\newline
- jealousy\newline
- education\newline
- fear\newline
- excitement\newline

I believe the correct choice is "excitement", here's why:\\
\hline
\textbf{Completion A} &  Someone seeing new presents at a birthday party means that there is new and exciting things happening. This is exciting for everyone at the party, especially the birthday person, because they are likely to receive new toys that they have not seen before. New toys mean new possibilities, and this is exciting for everyone at the party because they can imagine what this new possibility could be.\\
\hline
\textbf{Completion B} &  New presents at a birthday party can be a source of excitement for some people, while for others it can be a source of fear, jealousy or even envy.\\
\hline
\end{tabularx}
\end{table}

\begin{table}
    \centering
    \footnotesize
    \caption{Comparison of \textit{dolly-v2-7b} vs. \textit{dolly-v2-12b}}
    \begin{tabularx}{\textwidth}{|l|X|}
        \hline
        \rowcolor{gray!30}
        \multicolumn{2}{|l|}{Ranked by Cross-Entropy} \\
        \hline
\multicolumn{2}{|c|}{\textbf{\textsc{CommonGen} - Top 20\%}} \\
\hline
\hline
\textbf{Prompt} & Put the concepts together to form a sentence: raft, pull, water.\\
\hline
\textbf{Completion A} & Rafts are pulled by water.\\
\hline
\textbf{Completion B} & Rafting is a fun form of outdoor recreation that combines the thrill of rafting with the beauty of nature. Rafting involves floating down a river in a rubber raft, with guides pulling you by rope over whitewater. When done correctly, rafting can be quite relaxing, as you are pulled over gentle riffles by the current, with the peacefulness interrupted only by the sound of the river and the crack of the guides rope. Sometimes people call out "rapids" when they are in a whitewater rafting trip, and these are rapid-fire words that the guides yell out to help the rafters know what to expect in the river. Whitewater rafting is also available in the form of a "Class III+" trip, which means the rapids are more intense than those in a "Class II" trip, and are suitable for expert rafters only. \\
\hline
\multicolumn{2}{|c|}{\textbf{\textsc{CommonGen} - Bottom 20\%}} \\
\hline
\textbf{Prompt} &  Put the concepts together to form a sentence: groom, cake, cut, bride.\\
\hline
\textbf{Completion A} & Groom and bride are both terms used to describe the head of a household. A groom will typically celebrate his wedding with a cake, but a bride will typically celebrate her wedding with a cutting of the cake. A cutting is a term used in Buddhism to describe the ceremony where monks receive alms food in return for teaching meditation lessons.\\
\hline
\textbf{Completion B} &  Grooming is a vital part of being a bride. It shows that you care for your appearance and are taking care of your own hygiene. It also shows that you are taking the time to prepare for your big day. A grooming kit is a great way to show your preparation and love of detail. A grooming kit can include things like hairbrushes, hair ties, hair spray, hair gel, hair scissors, etc. Grooming is also a symbol of new beginnings, and a bride should be excited for her big day. A grooming kit can be a great way to make the day special and prepare for the big celebration.\\
\hline
\end{tabularx}
\end{table}

\newpage
\subsection{Annotation Guidelines}
\label{sec:eval-guide}
We strive to maintain a consistent evaluation framework for our human feedback experiments conducted within our work.
The following guidelines have been established to assist the annotators in conducting systematic and standardized pairwise comparisons of model completions.

\hspace{0.5cm}
\begin{mdframed}[linewidth=.5pt]
As an annotator, you have the exciting task of selecting the best completion by evaluating how well each one covers the given guidelines in relation to the other completion. Take a thorough look at both completions, and the winner should be the one that truly stands out compared to its counterpart. Remember, it's not necessary to check off all rules perfectly; instead, consider which completion adheres to the guidelines to the highest extent. If both completions demonstrate a similar level of adherence, feel free to choose the `Both good' option.
On the other hand, if neither of the options can fulfill the task or adhere to these rules, feel free to select the `Both bad` option. \textbf{Rules} are:
\setlist[enumerate]{nosep}
\begin{enumerate}
\item  \textbf{Task fulfilment}: The most important rule is whether completions fulfill the task. For example in case of \textit{`common\_gen\_Put\_together`} prompts, all concept words should be included in one, and only one, sentence as stated in the prompt. Completions that separate the content into multiple sentences might not be in line with the original task. However, if a completion has more than one sentence, and one of those sentences fulfills the task requirement, it is also considered acceptable.
\item \textbf{Grammar}: If both completions respect the first rule, grammar would be the next important deciding factor.
\item \textbf{Semantic}: If the sentence makes sense at all should be considered.
\item \textbf{Creativity}: Personal preference can play a part in your evaluation process when other criteria are fulfilled.
\end{enumerate}

\end{mdframed}

\subsection{Dataset-Specific Instructions}
\label{sec:eval-instruct}

This section provides instructions tailored to each dataset, complementing the main annotation guidelines to further ensure consistent assessments.

\hspace{0.5cm}
\begin{mdframed}[linewidth=.5pt]

\noindent\textbf{\textsc{Soda}} --- A multi-turn conversation with some context in the intro truncated at a random next turn utterance:
\vspace{-1em}
    \begin{itemize}
        \item No specific task is explicitly described
        \item The completion is expected to include the next turn in the conversation, preferably by incorporating the given context
        \item A completion may also include the whole prompt (context and/or dialogue), then the next turn of the conversation. It can also extend to multiple turns completion. As long as the dialogue is fluent and within context, this is also acceptable.
    \end{itemize}
    \vspace{-1em}
\noindent\textbf{\textsc{AdversarialQA}} --- Information extractive task given some context and a related question:
\vspace{-1em}
    \begin{itemize}
        \item The completion should include information from given context.
        \item The completion can be a one word answer or a whole paragraph.
        \item The labels for this dataset can be found shared if needed for reference to speed up the annotation process, as some questions tend to be ``tricky".
    \end{itemize}
    \vspace{-1em}
\noindent\textbf{\textsc{CommonGen}} --- Construct a sentence with all provided concept words:
\vspace{-1em}
    \begin{itemize}
        \item Concept words examples: sit —> sat, ball —> football, swim —> swimmer
        \item A completion should include at least one sentence with all concept words.
        \item A completion may start with “The sentence is”
        \item A completion may also include a description of the example sentence.
        \item A completion with concept words split between two or more sentences is not Ok.
    \end{itemize}
    \vspace{-1em}
\noindent\textbf{\textsc{CommonsenseQA}} --- Explain why one choice among others is the correct answer:
\vspace{-1em}
    \begin{itemize}
        \item  Completions may include the prompt itself or part of it. This is Ok.
        \item  The completion should have a decent level of common sense, logical or moral reasoning, depending on the situation.
        \item Some completions will start hallucinating scenarios as a possible explanation, these are per-case acceptable. If you are unsure, please share the example on the slack channel!
    \end{itemize}
    \vspace{-1em}

\end{mdframed}

\section{Pairwise Comparison Results}
\label{sec:app_rates}

\begin{figure}[thb]
    \centering
    \begin{subfigure}{0.45\textwidth}
        \centering
        \includegraphics[width=\linewidth]{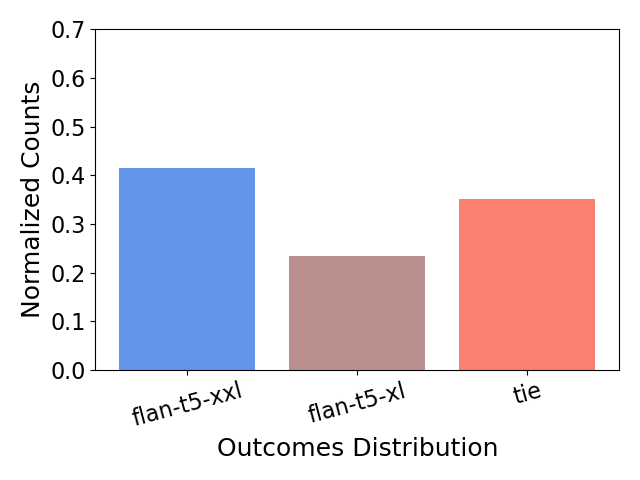}
        \caption{flan-t5-xxl vs. flan-t5-xl}
        \label{fig:flans-win-rates}
    \end{subfigure}
    \hfill
    \begin{subfigure}{0.45\textwidth}
        \centering
        \includegraphics[width=\linewidth]{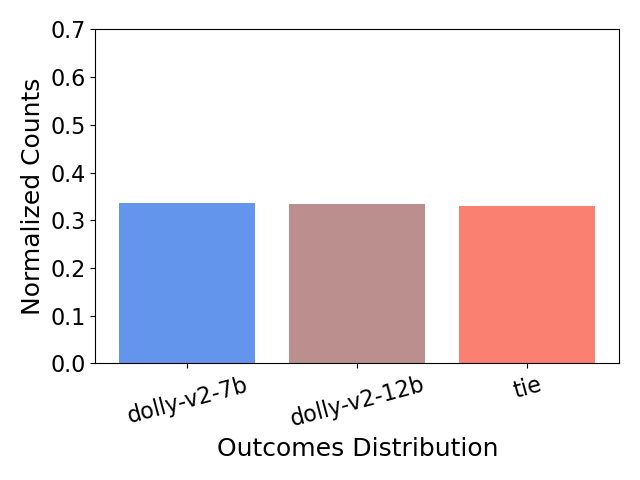}
        \caption{dolly-v2-12b vs. dolly-v2-7b}
        \label{fig:dollys-win-rates}
    \end{subfigure}
    \\
    \begin{subfigure}{0.45\textwidth}
        \centering
        \includegraphics[width=\linewidth]{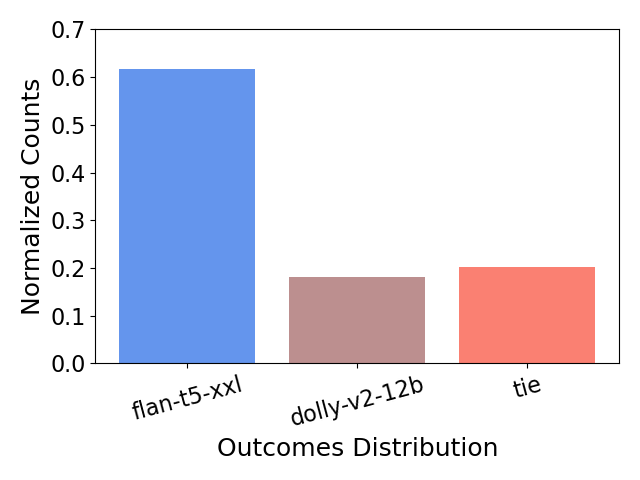}
        \caption{flan-t5-xxl vs. dolly-v2-12b}
        \label{fig:flan-dolly-win-rates}
    \end{subfigure}
    \hfill
    \begin{subfigure}{0.45\textwidth}
        \centering
        \includegraphics[width=\linewidth]{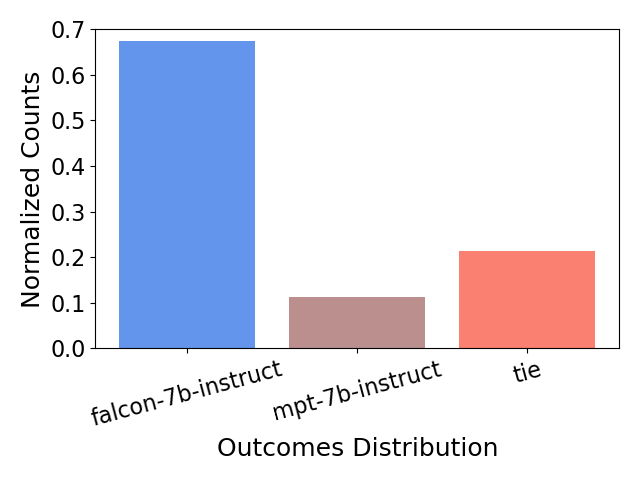}
        \caption{MPT-7b-instruct vs. falcon-7b-instruct}
        \label{fig:mpt-falcon-win-rates}
    \end{subfigure}
    \caption{Barplots representing the win rates per model for each of our four experiments, alongside the rate of tie outcomes.}
    \label{fig:win-rates-all}
\end{figure}

We detail the outcome rates of each pairwise comparison discussed in Section \ref{sec:exp}.
The summarized results are depicted in Figure~\ref{fig:win-rates-all}.
We observe that models within the same family, particularly those of proximate number or parameters, such as \textit{dolly-v2-12b} vs. \textit{dolly-v2-7b}, display analogous responses attributed to their shared architectures and overlapping training data, resulting in higher tie rates.
On the other hand, comparisons between models from different families yield lower tie rates.
Such models exhibit varied writing styles, stemming from their unique training backgrounds, making evaluations more decisive.
Our evaluation prompts set, curated to encompass diverse tasks, accentuates these output disparities, with certain tasks proving more intuitive for some models and challenging for others.

\section{Elo Scores Analysis Results}
\label{sec:app_elo}

\begin{figure}[htb]
    \centering

    \begin{subfigure}[t]{0.80\textwidth} 
        \includegraphics[width=\textwidth]{images/flan_vs_flan/dots_plot_elo_scores_flan_vs_flan.png}
        \subcaption{flan-t5-xxl vs. flan-t5-xl}
    \end{subfigure}

    \vspace{1em}%

    \begin{subfigure}[t]{0.80\textwidth} 
        \includegraphics[width=\textwidth]{images/dolly_vs_dolly/dots_plot_elo_scores_dolly_vs_dolly.png}
        \subcaption{dolly-v2-7b vs. dolly-v2-12b}
    \end{subfigure}

    \vspace{1em}%
    
    \begin{subfigure}[t]{0.80\textwidth} 
        \includegraphics[width=\textwidth]{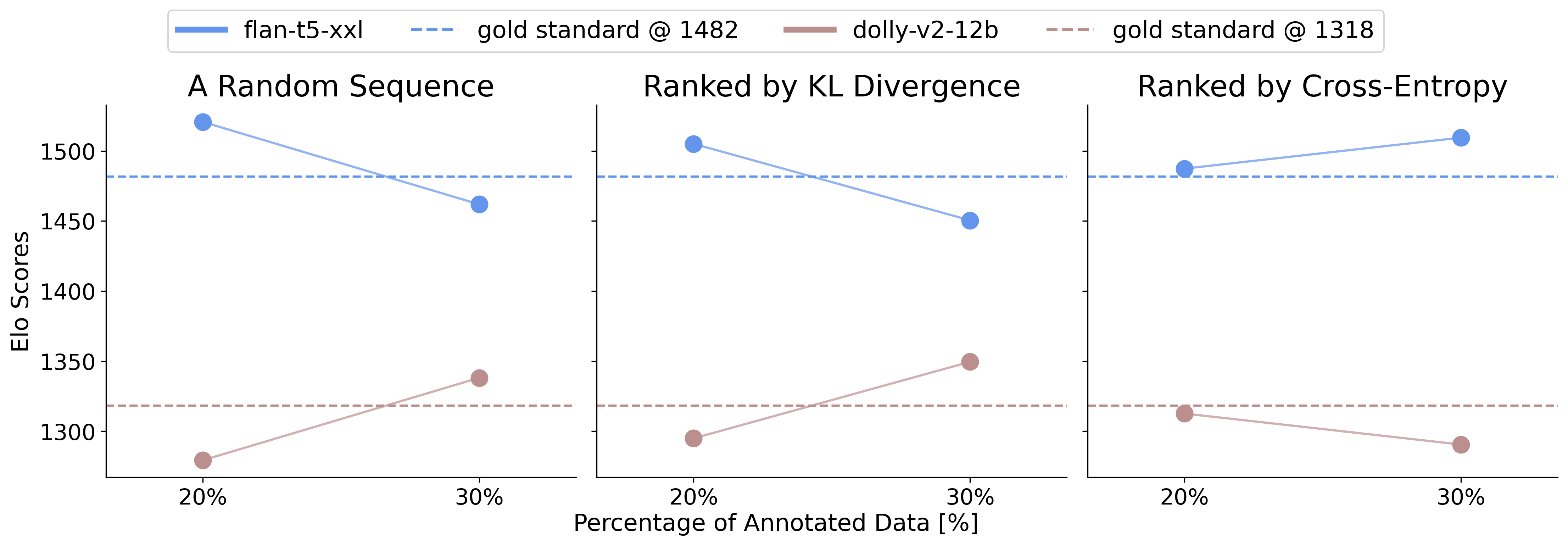}
        \subcaption{flan-t5-xxl vs. dolly-v2-12b}
    \end{subfigure}

    \vspace{1em}%

    \begin{subfigure}[t]{0.80\textwidth} 
        \includegraphics[width=\textwidth]{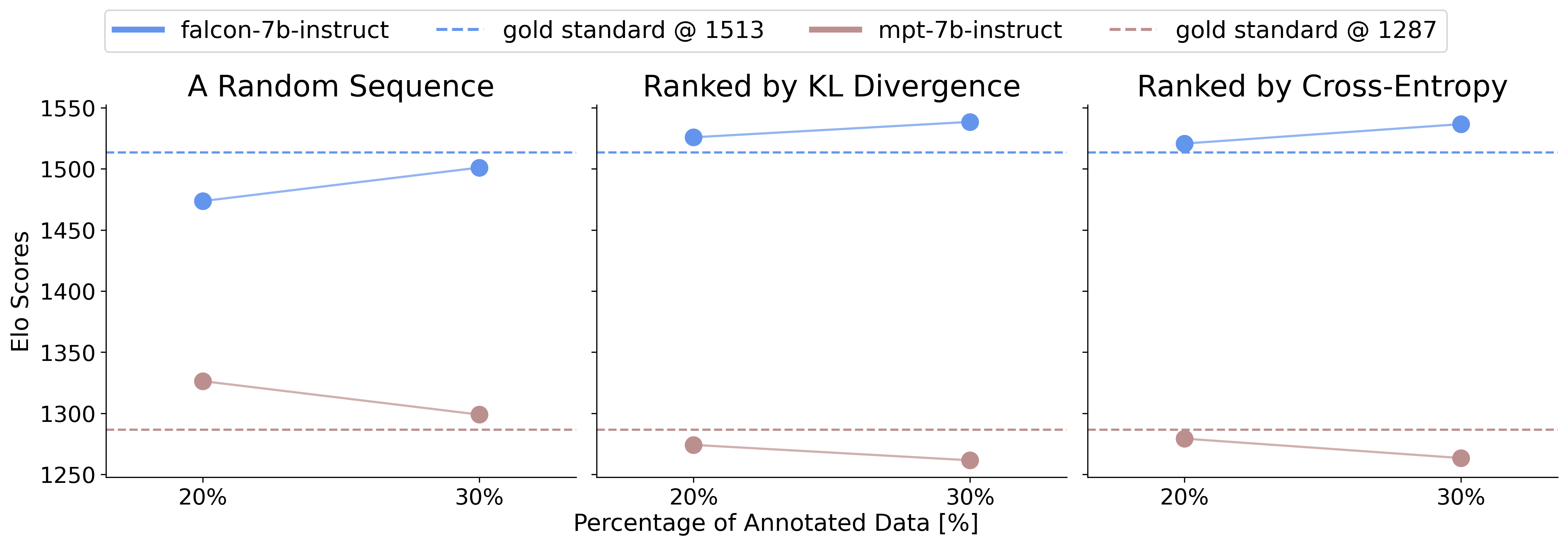}
        \subcaption{MPT-7b-instruct vs. falcon-7b-instruct}
    \end{subfigure}

    \caption{\textbf{Elo Scores} for inter- and intra-family experiments. Dashed lines indicate the `Gold Standard' values for each model; refer to Table \ref{tab:elo_scores} for details.}
    \label{fig:elo-robust-all}
\end{figure}

\begin{figure}[htb]
    \centering
    \begin{subfigure}[t]{0.90\textwidth} 
        \includegraphics[width=\textwidth]{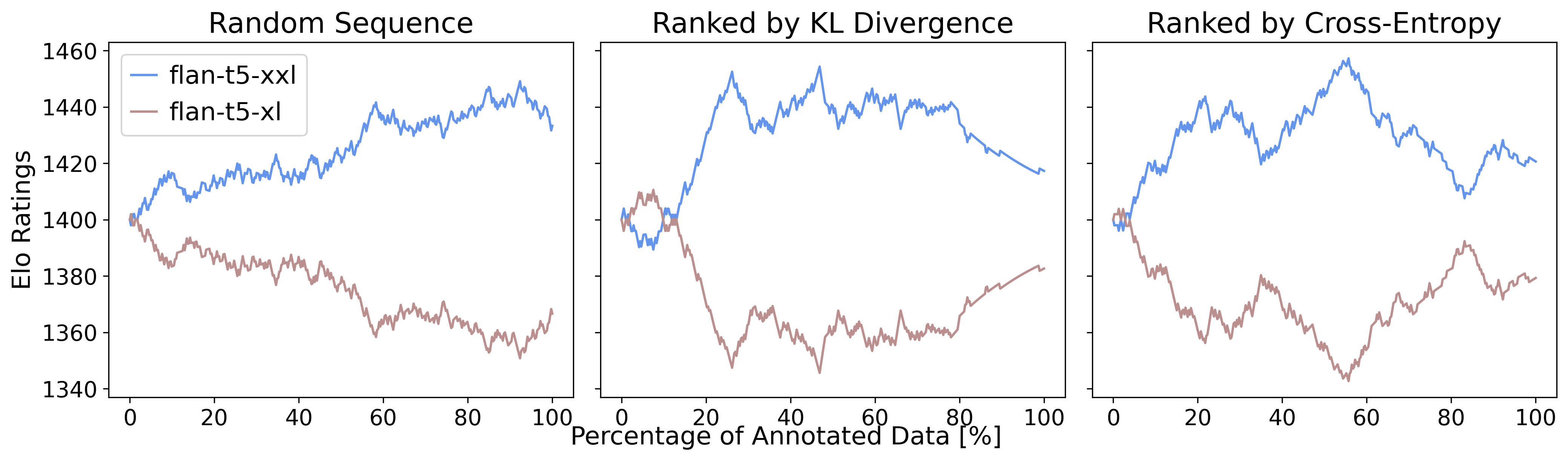}
        \subcaption{flan-t5-xxl vs. flan-t5-xl}
    \end{subfigure}
    \vspace{1em}%
    \begin{subfigure}[t]{0.90\textwidth} 
        \includegraphics[width=\textwidth]{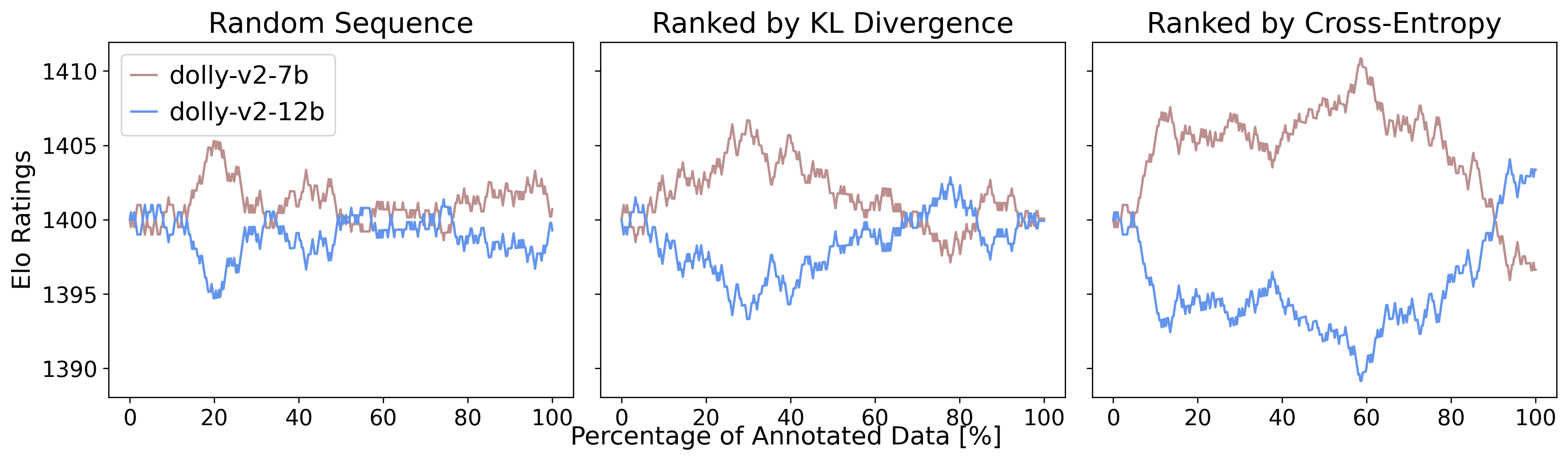}
        \subcaption{dolly-v2-7b vs. dolly-v2-12b}
    \end{subfigure}
    \vspace{1em}%
    \begin{subfigure}[t]{0.90\textwidth} 
        \includegraphics[width=\textwidth]{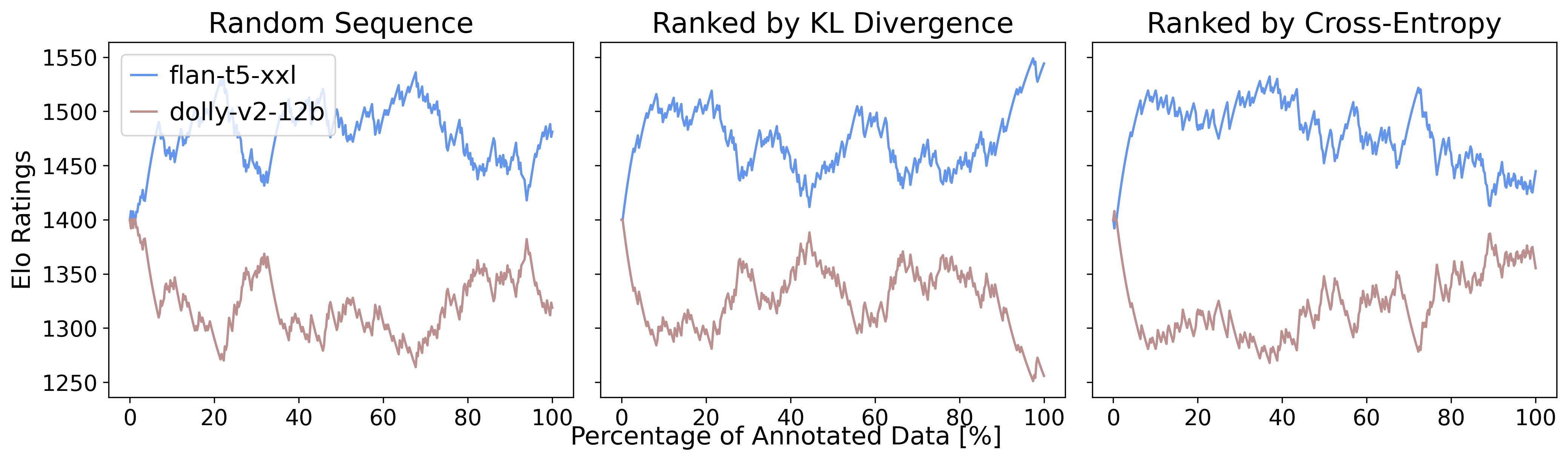}
        \subcaption{flan-t5-xxl vs. dolly-v2-12b}
    \end{subfigure}
    \vspace{1em}%
    \begin{subfigure}[t]{0.90\textwidth} 
        \includegraphics[width=\textwidth]{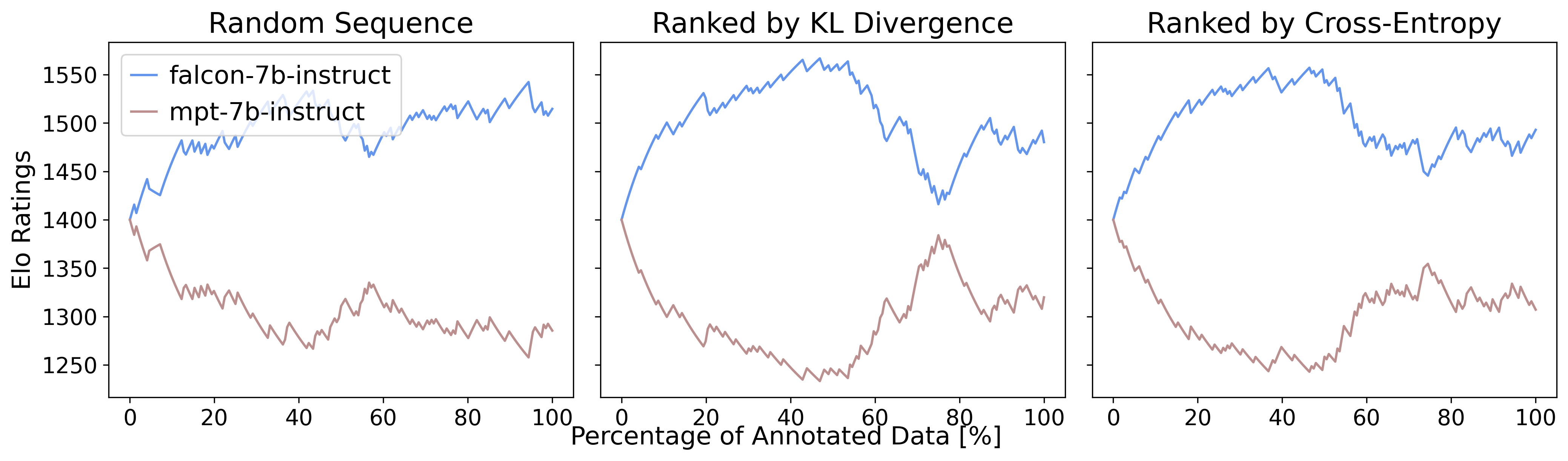}
        \subcaption{MPT-7b-instruct vs. falcon-7b-instruct}
    \end{subfigure}
     \caption{\textbf{Elo Ratings Analysis} for Inter- and Intra-Family Model Comparisons. We contrast the ranking methods, KL Divergence and Cross-Entropy, against a Random Sequence strategy to demonstrate their impact on ratings update per evaluation instance.}
    \label{fig:elo-converge-all}
\end{figure}

Building upon section~\ref{sec:elo}, where we examine the impact of our ranking metrics on the Robustness of Elo Scores used to relatively rank models, we provide a direct comparison between inter- and intra-family results (see Fig.~\ref{fig:elo-robust-all}).
Additionally, we chart the progression of Elo ratings throughout the evaluation process.
Compared to a random sequence (refer to  Fig.~\ref{fig:elo-converge-all}), our ranking metrics amplify the divergence of Elo scores, especially within the early top k\% of evaluation instances, up to top 40 percentile.

